%% file: root.tex
\documentclass[letterpaper]{article} %
\usepackage{aaai24}  %
\usepackage{times}  %
\usepackage{helvet}  %
\usepackage{courier}  %
\usepackage[hyphens]{url}  %
\usepackage{graphicx} %
\usepackage{natbib}  %
\usepackage{caption} %
\usepackage[utf8]{inputenc} %
\usepackage{booktabs}       %
\usepackage{amsfonts}       %
\usepackage{nicefrac}       %
\usepackage{microtype}      %
\usepackage{xcolor}         %

\usepackage{amsmath}
\usepackage{amssymb}
\usepackage{mathtools}
\usepackage{amsthm}
\usepackage{subcaption}

\theoremstyle{plain}

\theoremstyle{definition}

\theoremstyle{remark}

\DeclareMathOperator*{\argmax}{arg\,max}
\newcommand{\norm}[1]{\left\lVert#1\right\rVert_2}

\usepackage{algorithm}
\usepackage[noend]{algorithmic}
\newtheorem{thm}{Theorem}[subsection]%

\definecolor{green}{RGB}{11,155,13}

\newcommand{\colorphi}{\textcolor{green}{\phi}}
\newcommand{\colorpsi}{\textcolor{orange}{\psi}}

\definecolor{purple}{RGB}{102, 0, 204}
\definecolor{brown}{RGB}{153, 76, 0}
\definecolor{grey}{RGB}{102,102,102}
\definecolor{causal_blue}{RGB}{255, 128, 0}
\definecolor{causal_green}{RGB}{0, 153, 0}
\definecolor{reward_orange}{RGB}{215,155,0}
\definecolor{reward_green}{RGB}{0, 153, 0}
\definecolor{cdl_yellow}{RGB}{214,182,86}
\definecolor{cbm_green}{RGB}{0, 153, 0}

\usepackage{dsfont}

\usepackage{listings}
\definecolor{codegreen}{rgb}{0,0.6,0}
\definecolor{codegray}{rgb}{0.5,0.5,0.5}
\definecolor{codepurple}{rgb}{0.58,0,0.82}
\definecolor{backcolour}{rgb}{0.95,0.95,0.92}

\lstdefinestyle{mystyle}{
    backgroundcolor=\color{backcolour},   
    commentstyle=\color{codegreen},
    keywordstyle=\color{magenta},
    numberstyle=\tiny\color{codegray},
    stringstyle=\color{codepurple},
    basicstyle=\ttfamily\footnotesize,
    breakatwhitespace=false,         
    breaklines=true,                 
    captionpos=b,                    
    keepspaces=true,                 
    numbers=left,                    
    numbersep=5pt,                  
    showspaces=false,                
    showstringspaces=false,
    showtabs=false,                  
    tabsize=2
}
\usepackage{diagbox}
\usepackage{makecell}
\usepackage{array, arydshln}
\usepackage{multirow}
\makeatletter
\patchcmd{\@makecaption}
  {\scshape}
  {}
  {}
  {}
\makeatother

\title{Building Minimal and Reusable\\Causal State Abstractions for Reinforcement Learning}

\author {
    Zizhao Wang\equalcontrib\textsuperscript{\rm 1},
    Caroline Wang\equalcontrib\textsuperscript{\rm 1},
    Xuesu Xiao\textsuperscript{\rm 2},
    Yuke Zhu\textsuperscript{\rm 1},
    Peter Stone\textsuperscript{\rm 1, \rm 3}
}
\affiliations {
    \textsuperscript{\rm 1}The University of Texas at Austin,
    \textsuperscript{\rm 2}George Mason University,
    \textsuperscript{\rm 3}Sony AI\\
    zizhao.wang@utexas.edu, caroline.l.wang@utexas.edu, xiao@gmu.edu, yukez@cs.utexas.edu, pstone@cs.utexas.edu
}

\begin{document}
\maketitle

\begin{abstract}
Two desiderata of reinforcement learning (RL) algorithms are the ability to learn from relatively little experience and the ability to learn policies that generalize to a range of problem specifications. 
In factored state spaces, one approach towards achieving both goals is to learn state abstractions, which only keep the necessary variables for learning the tasks at hand. 
This paper introduces \textit{Causal Bisimulation Modeling} (CBM), a method that learns the causal relationships in the dynamics and reward functions for each task to derive a minimal, task-specific abstraction. 
CBM leverages and improves implicit modeling to train a high-fidelity causal dynamics model that can be reused for all tasks in the same environment. 
Empirical validation on manipulation environments and Deepmind Control Suite reveals that CBM's learned implicit dynamics models identify the underlying causal relationships and state abstractions more accurately than explicit ones. Furthermore, the derived state abstractions allow a task learner to achieve near-oracle levels of sample efficiency and outperform baselines on all tasks. 
\end{abstract}

\input{contents/01_intro}
\input{contents/02_related}
\input{contents/03_background}

\input{contents/04_approach_part1}

\input{contents/05_approach_part2}
\input{contents/06_experiments}
\input{contents/07_conclusions}

\newpage
\section*{Acknowledgements}
This work has taken place in the Learning Agents Research
Group (LARG) at UT Austin.  LARG research is supported in part by NSF
(FAIN-2019844, NRT-2125858), ONR (N00014-18-2243), ARO (E2061621),
Bosch, Lockheed Martin, and UT Austin's Good Systems grand challenge.
Peter Stone serves as the Executive Director of Sony AI America and
receives financial compensation for this work.  The terms of this
arrangement have been reviewed and approved by the University of Texas
at Austin in accordance with its policy on objectivity in research.

\bibliography{references}
\newpage
\appendix

\input{contents/08_appendix}

\end{document}

%% file: contents/01_intro.tex
\section{Introduction} 
\label{sec:intro}

\begin{figure}[ht!]
  \centering
  \includegraphics[width=0.85\columnwidth]{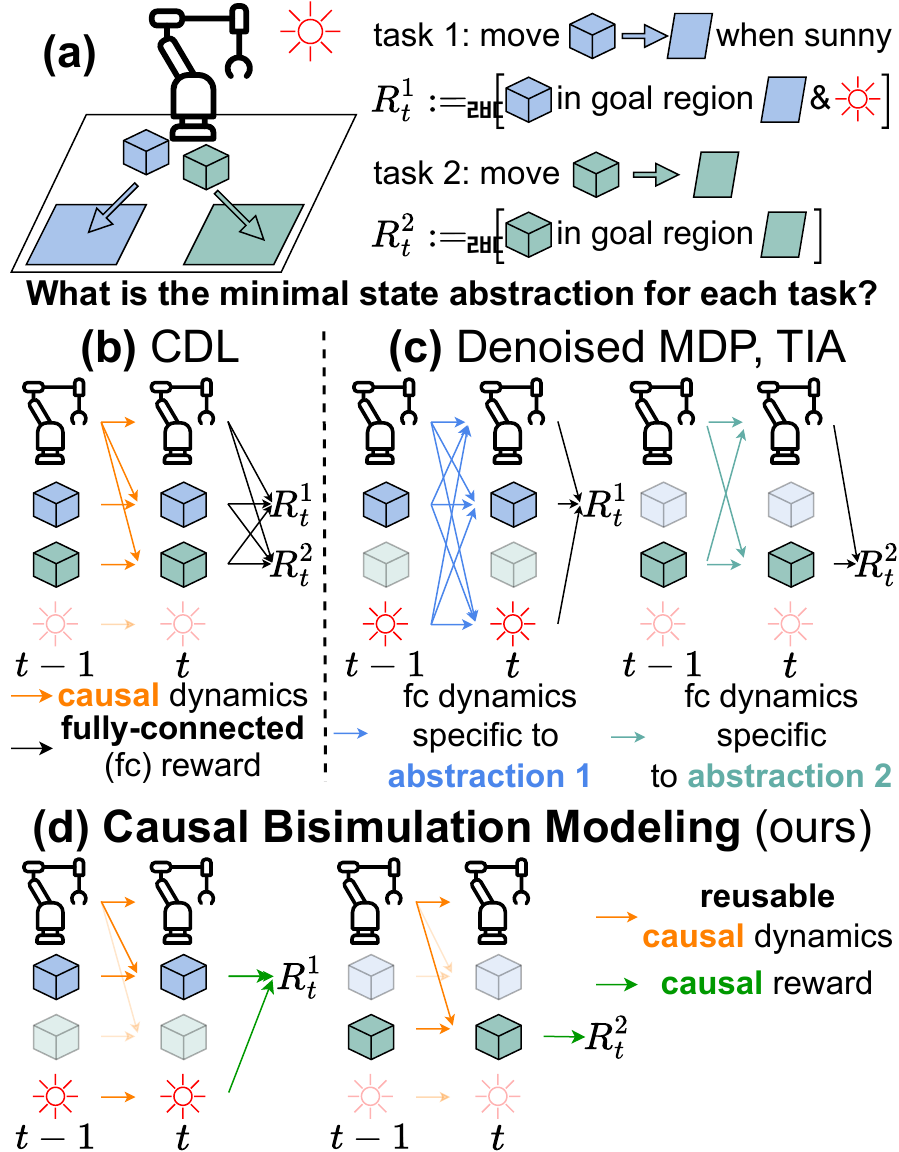}
  \caption{\small \textbf{(a)} Two tasks are defined by rewards $R^1, R^2$, and consist respectively of moving the blue and green blocks to their goal regions. Task 1 additionally requires moving the block only when it is sunny. Variables that are ignored by a state abstraction are semi-transparent. \textbf{(b)} CDL \citep{wang2022cdlicml} learns causal dependencies in the dynamics, but its derived state abstraction keeps \textit{all} controllable state variables and ignores action-irrelevant ones, and thus the abstraction is non-minimal for task 2 and cannot learn task 1 due to its omission of the sun. \textbf{(c)} TIA and Denoised MDP \citep{fu2021learning, pmlr-v162-wang22c} can learn more concise abstractions (minimal in this example), but they require training fully-connected dynamics from scratch for each task. \textbf{(d)} In addition to the implicit \textcolor{causal_blue}{\emph{causal dynamics}} that can be \textcolor{causal_blue}{\emph{reused}} for all tasks, CBM identifies which variables affect the reward and derives a minimal state abstraction from the \textcolor{causal_green}{\emph{causal reward}} models.}
  \label{fig:intro}
  \vspace{-15pt}
\end{figure}

Reinforcement learning (RL) is a general paradigm that enables autonomous decision-making in unknown environments. A common deficiency of deep RL algorithms is their sample inefficiency and lack of generalization to unseen states, thus limiting their applicability in data-expensive or safety-critical tasks. One way to improve sample efficiency and generalization is to learn a \textit{state abstraction} which reduces the task learning space and eliminates unnecessary information that may lead to spurious correlations.
Determining the optimal abstraction necessitates understanding which state variables affect the reward and how those variables are influenced by others during state transitions.

Prior methods obtain the desired abstractions by searching for the smallest subset of state variables that can predict the reward accurately while ensuring the subset is self-predictable in dynamics. Yet, their \textit{dense} dynamics models are specific to the subset and thus have to be learned from scratch for each new task \citep{fu2021learning, pmlr-v162-wang22c, zhang2020learning}, as shown in Fig. \ref{fig:intro}. Such approaches overlook a key characteristic of realistic problems: we often wish to build agents that solve multiple instances of tasks in the same environment, e.g. different cooking skills in a kitchen. To learn multi-task dynamics models, recent works seek to learn \textit{causal} dynamical dependencies between state variables, from which they derive a task-\textit{independent} abstraction \citep{ding2022generalizing, wang2021task, wang2022cdlicml}. A notable example is Causal Dynamics Learning (CDL)~\cite{wang2022cdlicml} which identifies all state variables that can be affected by the action and retains them in the abstraction.

While dynamics-based state abstractions only need to be learned once for the environment and can then be applied to any downstream task, we observe three weaknesses. 
First, the abstraction may not be minimal for a significant number of downstream tasks, since many tasks require manipulating only a subset of controllable variables. In such cases, CDL's abstraction could be further reduced to improve sample efficiency and generalization, as shown in Fig. \ref{fig:intro}(b). 
Second, ignoring state variables that are action-irrelevant limits CDL's application to many tasks. For example, for an autonomous vehicle, CDL’s abstraction will ignore the traffic light, as the traffic light cannot be affected by the vehicle. In such cases, CDL’s abstraction will cause the vehicle to be unable to follow traffic rules.
Third, CDL employs explicit modeling of dynamics, directly predicting the next state as $\hat{s}_{t+1} = f(s_t, a_t)$ where $f$ is a generic function parameterized by neural networks. 
However, prior works have shown \textit{implicit modeling} ($\hat{s}_{t+1}=\argmax_{s_{t+1}\in \mathcal{S}} g(s^i_{t+1}; s_t, a_t)$ where $g$ is a critic function, see Sec. \ref{subsec:causal_learning}) generally achieves higher accuracy in model learning, particularly for non-smooth dynamics in real-world physical systems \citep{florence2022implicit, Song2021HowTT}. 
For instance, in robot manipulation, the object cannot be moved by the robot until they are in contact. 
In such environments, we show that inaccuracies of explicit modeling will lead to incorrect dynamical dependencies and thus non-minimal or incorrect state abstractions.

To address these weaknesses of CDL, we introduce Causal Bisimulation Modeling (CBM), a method that (1) learns \textbf{shared} task-agnostic dynamics between tasks while recovering a minimal, \textbf{task-specific} state abstraction, and (2) models causal dynamics dependencies with implicit models. 
Regarding the first contribution, in addition to dynamical relationships, CBM further infers which state variables affect the reward function with a causal reward model. In this way, CBM identifies state variables relevant to each task to further refine the state abstractions.
The resultant causal abstraction is equivalent to bisimulation, a minimal abstraction that preserves the optimal value \citep{ferns2011bisimulation}. 
Regarding the second contribution, to the best of our knowledge, CBM is the first work that recovers causal dependencies with implicit models. To this end, CBM identifies and addresses two key problems of estimating conditional mutual information (CMI) with implicit models, allowing them to surpass explicit ones in both predictive accuracy and the identification of causal dependencies.

We validate CBM in robotic manipulation and Deepmind Control Suite, showing that (1) implicit models learn dynamical relationships and state abstractions more accurately compared to the explicit ones, and (2) CBM's task-specific state abstractions significantly improve sample efficiency and generalization compared to task-independent ones.

%% file: contents/02_related.tex
\section{Related Work}
\label{sec:related}

\paragraph{Model-based State Abstractions for Decision Making}

Learned dynamics and reward models can be used in various ways for downstream task learning. 
Some methods directly use the learned models for planning \citep{williams2017information, chua2018deep, nagabandi2018neural} or generate synthetic rollouts for reinforcement learning \citep{kurutach2018model, Janner2019WhenTT}. 
Others use learned models to improve $Q$-value estimates \citep{feinberg2018model, amos2021model}, or generate state abstractions \citep{li2006towards, fu2021learning, pmlr-v162-wang22c, wang2021task, zhang2019learning}. 
This work belongs to the last class of methods. 

The work closest to our CBM is \citet{wang2022cdlicml} (CDL), which also learns a causal dynamics model and then derives a state abstraction for downstream task learning. As discussed in the introduction, CDL's abstraction is not minimal because it does not consider task information. In contrast,  CBM considers causal reward relationships to derive a theoretically minimal, task-specific state abstraction. 

Among model-based methods that learn task-specific state abstractions, the most closely related works are TIA \citep{fu2021learning}, denoised MDP \citep{pmlr-v162-wang22c}, and ASR \citep{huang2022action}. 
Those methods learn dense, non-causal dynamics models from scratch for each task, which fails to take advantage of shared structures between the tasks. In contrast, CBM learns underlying causal dynamics that are shared between tasks in the same environment and applies the same dynamics model to all downstream tasks. 
\paragraph{Implicit Models for Dynamics Learning}
Implicit models \citep{Teh2003EnergyBasedMF, Welling2002ANL} have been widely used in many areas of machine learning, including image generation \citep{du2019langevinebm}, natural language processing \citep{Bakhtin21textebm, He2021JointEM}, and density estimation \citep{Saremi2018DeepEE, Song2019SlicedSM}. This is largely due to its ability to generalize probabilistic and deterministic approaches to classification, regression, and estimation \citep{lecun2006ebmtutorial, Song2021HowTT}. 

Implicit modeling approaches have also been applied to reinforcement learning and the closely related problem of imitation learning, for modeling policies \citep{florence2022implicit}, value functions \citep{Haarnoja2017ReinforcementLW, Haarnoja2018SoftAC}, and dynamics \citep{Pfrommer2020ContactNetsLO, Wang2020CLOUDCL}. Multiple works have noted that implicit approaches are better able to model discontinuous surfaces, which is particularly advantageous for modeling the discontinuous contact dynamics common in robotics \citep{Pfrommer2020ContactNetsLO, florence2022implicit}.
For such dynamics, though explicit models theoretically can capture such discontinuities through activation functions, empirically, they linearly interpolate between discontinuity boundaries when training data are finite, as found by \citet{florence2022implicit}. 

%% file: contents/03_background.tex
\vspace{-5pt}
\section{Background}

We formulate our problem with Markov Decision Processes and adopt key concepts from CDL \citep{wang2022cdlicml}.

\paragraph{Factored Markov Decision Processes}
We model $K$ tasks in the same environment as a set of factored Markov decision processes (MDPs), $\mathcal{M}^k = (\mathcal{S}, \mathcal{A}, \mathcal{T}, \mathcal{R}^k)$. These MDPs have the same (1) finite (bounded) state space, consisting of $d_\mathcal{S}$ state variables (factors) denoted as $\mathcal{S} = \mathcal{S}^1 \times \cdots \times \mathcal{S}^{d_\mathcal{S}}$, where each variable $\mathcal{S}^i$ is a scalar,
(2) $d_\mathcal{A}$-dimensional action space, denoted as $\mathcal{A} \subseteq \mathbb{R}^{d_\mathcal{A}}$, and
(3) transition probability $\mathcal{T}(s_{t+1} | s_t, a_t)$ (i.e., dynamics). 
However, each MDP has its own reward function $\mathcal{R}^k: \mathcal{S} \times \mathcal{A} \rightarrow \mathbb{R}$. 

Similar to prior model-based state abstraction learning methods, the goal of CBM is to learn the dynamics and reward functions from data, and to derive a task-specific abstraction for task learning. 
Following CDL, CBM assumes that the transitions of each state variable ${S}^i$ are independent, i.e., $\mathcal{T}$ can be decomposed as $\mathcal{T}(s_{t+1} | s_t, a_t) = \prod_{i=1}^{d_\mathcal{S}} p(s^i_{t+1} | s_t, a_t)$. 
Though the assumption does not hold for all variables---for instance, quaternion variables representing object rotations in the manipulation environments---in practice, we find that our method can still learn dynamics accurately in such environments. 
For simplicity, we use $x_t$ to denote all state variables and the action at $t$, i.e., $\boldsymbol{x_t = \{s^1_t, \cdots, s^{d_\mathcal{S}}_t, a_t\}}$ and $x^{\neg i}_t$ to denote all those variables except for $s^i_t$, i.e., $\boldsymbol{x^{\neg i}_t = x_t \setminus \{s^i_t\}}$.

\paragraph{Causal Dynamics Learning (CDL)}
Instead of using a dense model, CDL models the dynamics as a causal graphical model \cite{pearl2009causality} and recovers the necessary dependencies between each state variable pair $(\mathcal{S}^i_t, \mathcal{S}^j_{t+1})$ as well as $(\mathcal{A}_t, \mathcal{S}^j_{t+1})$ using causal discovery methods \citep{mastakouri2021necessary}. 
Then, aiming at improving sample efficiency during task learning, CDL derives a task-independent state abstraction by keeping (1) \textbf{controllable} state variables, i.e., those that can be changed by the action directly or indirectly, and (2) \textbf{action-relevant} state variables, i.e., those that cannot be changed by the action but can affect the action's influence on controllable variables. 
However, this abstraction keeps all controllable variables, while many tasks only need the agent to control one or a few of them, suggesting some variables in CDL's abstraction may be redundant.

%% file: contents/04_approach_part1.tex
\section{Causal Bisimulation Modeling (CBM)}
\label{sec:method}
This section describes two main contributions of CBM: obtaining a task-specific state abstraction by augmenting causal dynamics models with causal reward modeling, and recovering accurate causal dynamics when using implicit models.

\subsection{Causal Reward Model for Task-Specific Abstraction}
\label{subsec:causal_reward_learning}

\begin{figure}[t]
    \centering
    \includegraphics[width=0.9\columnwidth]{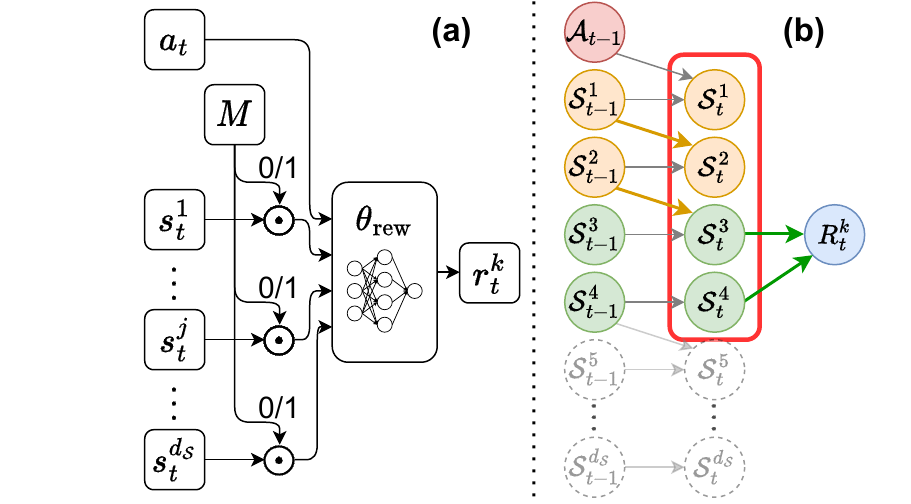}
    \vspace{-10pt}
    \caption{\small \textbf{(a)} The reward predictor architecture which can represent $p(r^k_t|a_t, M\odot s_t)$ conditioning any subsets of inputs by setting the binary mask $M$.
    \textbf{(b)} After CBM identifies the causal dependencies in dynamics and reward, its minimal state abstraction (marked by the red box) consists of (1) \textcolor{causal_green}{green} variables that affect the reward, and (2) \textcolor{reward_orange}{orange} variables that influence \textcolor{causal_green}{green} ones via dynamics.
    The semi-transparent state variables are ignored by the abstraction.}
    \label{fig:causal_illustration}
    \vspace{-10pt}
\end{figure}

Though CDL recovers the causal relationships in dynamics, it still uses a dense model for reward learning. Consequently, without knowing which state variables are causal parents of the reward (i.e., reward-relevant), there is no direct way to remove irrelevant variables to improve sample efficiency.

To resolve these issues, CBM learns a causal reward model following a similar strategy to what CDL uses to learn dynamics. 
Assuming that all state variables affect the reward for task $\mathcal{R}_t^k$ independently  (for notational simplicity, we omit the environment index $k$ for the reminder of the paper) and there are no dependencies between state variables at the same timestep, CBM examines the causal relationship between the state variable $\mathcal{S}_t^j$ and the reward $\mathcal{R}$ by learning two predictive models for the reward: (1) $p(r_t | x_t)$, which uses all variables for prediction, and (2) $p(r_t | x^{\neg j}_t)$, which ignores $s^j_t$ when predicting. 
Intuitively, if the prediction performance when ignoring $s^j_t$ is significantly lower than when including it, then the causal dependency $\mathcal{S}_t^j \rightarrow \mathcal{R}$ exists. 
More precisely, CBM evaluates whether the Conditional Mutual Information (\textsc{cmi}) is larger than a predefined threshold, i.e., 
$\textsc{cmi}^{jk} := \mathop{\mathbb{E}}\limits_{s_t,a_t,r_t}\left[\log\frac{p(r_t | x_t)}{p(r_t | x^{\neg j}_t)}\right] \ge \epsilon.$

As shown in Fig. \ref{fig:causal_illustration} (a), to make the method scalable, rather than training $p(r_t | x^{\neg j}_t)$ for each $j \in \{1, \dots, d_\mathcal{S}\}$, CBM combines all predictive models into one network $p_{\theta_\text{rew}}(r_t | M \odot x_t)$; where $\theta_\text{rew}$ is the network parameters and $M$ is a manually defined binary mask used to ignore some input variables when predicting $r_t$. During training, CBM maximizes the prediction likelihood of both $p_{\theta_\text{rew}}(r_t | x_t)$ where $M$ uses all inputs (i.e., all entries are set to 1), and $p_{\theta_\text{rew}}(r_t | x^{\neg j}_t)$ where the entry for $s^j_t$ is set to zero in $M$.

After recovering causal parents of $\mathcal{R}$, CBM derives a bisimulation by combining the causal reward model with the causal dynamics model, following the theorem below (see the Appendix for an explanation of how our setting satisfies the theorem's assumptions):
\begin{thm}[Connecting Bisimulation to Causal Feature Set (Thm 1 in \citet{zhang2020invariant})]
\label{theorem:bisimulation}
Consider an MDP $\mathcal{M}$ that satisfies Assumptions 1-3 in \citet{zhang2020invariant}. Let ${\textbf{P}_{\mathcal{R}}} \subseteq {1, . . . , d_\mathcal{S}}$ be the set of variables such that the reward $\mathcal{R}(s, a)$ is a function only of $s^{\textbf{P}_{\mathcal{R}}}$ ($s$ restricted to the indices in ${\textbf{P}_{\mathcal{R}}}$). Let ${\textbf{A}_{\mathcal{R}}}$ denote the ancestors of $\textbf{P}_{\mathcal{R}}$ in the causal graph corresponding to the transition dynamics of $\mathcal{M}$. Then the state abstraction $\phi(s)=s^{\textbf{A}_{\mathcal{R}}}$ is a bisimulation abstraction for reward $\mathcal{R}$.
\end{thm}

As illustrated in Fig. \ref{fig:causal_illustration} (b), CBM's abstraction is selected as the union of (1) all $\mathcal{S}^j$ that $\mathcal{R}$ depends on, i.e., $\mathcal{S}^{\textbf{P}_{\mathcal{R}}}$, and (2) all other state variables that can affect $\mathcal{S}^{\textbf{P}_{\mathcal{R}}}$ via dynamics and not already included in $\mathcal{S}^{\textbf{P}_{\mathcal{R}}}$. In other words, this union corresponds to $\mathcal{R}$'s causal ancestors (i.e., $\mathcal{S}^{\textbf{A}_{\mathcal{R}}}$) in the learned causal dynamics and reward graph, and thus being equivalent to bisimulation --- the \textit{minimal state abstraction} that preserves the optimal values \citep{dean1997model, ferns2011bisimulation}.

%% file: contents/05_approach_part2.tex
\subsection{Causal Discovery with Implicit Dynamics Models}
\label{subsec:causal_learning}

Implicit models have been shown to learn dynamics more accurately than explicit models \citep{florence2022implicit}.
Motivated by this finding, the goal of CBM's dynamics learning module is to recover causal relationships between states and action variables via an \textit{implicit} modeling approach. 
As in CBM's reward-learning approach, causal dependencies will also be detected by measuring the conditional mutual information $\textsc{cmi}^{ij}$ between $s^j_t$ and $s^i_{t+1}$ for each $i, j$ pair.
In this section, first, we introduce the implicit dynamics model. Second, we describe how CBM estimates CMI from the implicit dynamics with a prior method by \citet{sordoni2021decomposed}. Third, we discuss two CMI overestimation issues of the prior method and how CBM solves them.

\subsubsection{Implicit Dynamics Models}
\label{subsubsec:implicit_dynamics}
For the transition of each state variable in $\mathcal{S}^i$, we would like an implicit model $g^i(s^i_{t+1}; x_t)$ such that $g^i:\mathcal{S}^i \times \mathcal{S} \times \mathcal{A} \rightarrow \mathbb{R}$ to assign a high score to the label $s^i_{t+1}$ from the ground truth distribution, and low values to other labels. During dynamics prediction, the model selects the $s_{t+1}$ that maximizes the total score over all state variables: $\hat{s}^i_{t+1}=\argmax_{s^i_{t+1}\in \mathcal{S}^i} \sum_{i=1}^{d_\mathcal{S}}g^i(s^i_{t+1}; x_t)$. For notational simplicity, in the remainder of Sec. \ref{subsec:causal_learning}, we omit the state variable index $i$ on $g$ as it is clear from the input variable $s^i_{t+1}$. 
An implicit dynamics model can be trained by minimizing the contrastive InfoNCE loss, 
\begin{align}
\mathcal{L}_\text{NCE}(g) &= -\log \frac{e^{g(s^i_{t+1}; x_t)}}{e^{g(s^i_{t+1}; x_t)} + \sum_{n=1}^N e^{g(\tilde{s}^{i, n}_{t+1}; x_t)}}.
\end{align}

Minimizing the InfoNCE loss requires the ground truth label $s^i_{t+1}$, as well as $N$ negative examples $\{\tilde{s}^{i,n}_{t+1}\}_{n=1}^N \sim p(s^i_{t+1})$ sampled the value range $S^i_{t+1}$.
This loss encourages $g$ to distinguish the label $s^i_{t+1}$ from negative samples, i.e., extract information about $s^i_{t+1}$ from $x_t$. 
The remainder of the paper uses $\mathcal{L}_\text{NCE}(F(x;y))$ to denote the InfoNCE loss that encourages a generic model $F$ to extract information about $x$ from $y$. 

\subsubsection{CMI Estimation with Implicit Dynamics Models}
\label{subsubsec:demi}

\paragraph{Observation 1} 
\citet{oord2018representation} show that, for a generic model $F(x, y)$ that minimizes $\mathcal{L}_\text{NCE}$, the minimized $\mathcal{L}_\text{NCE}$ approximates the mutual information between $x$ and $y$, i.e., $\mathbb{E}\left[\log(N+1)-\mathcal{L}_\text{NCE}(F(x;y))\right] \approx \mathbb{E}\left[\frac{p(y|x)}{p(y)}\right]=I(x;y)$.
Inspired by this, \citet{sordoni2021decomposed} propose to estimate $\textsc{cmi}^{ij}$ using a conditional implicit model: 
\begin{align}
\label{eq:cmi_implicit}
&\textsc{cmi}^{ij}=
\mathop{\mathbb{E}}\left[\log\frac{(N+1) e^{\colorphi(s^i_{t+1}; s_t^j \textcolor{green}{| x^{\neg j}_t})}}{e^{\colorphi(s^i_{t+1}; s_t^j \textcolor{green}{| x^{\neg j}_t})} + \sum_{n=1}^N e^{\colorphi(\tilde{s}^{i, n}_{t+1}; s_t^j \textcolor{green}{| x^{\neg j}_t})}}\right], \nonumber\\
&\text{where } \tilde{s}^{i, n}_{t+1} \sim \textcolor{blue}{p(s^i_{t+1} | x^{\neg j}_t)}.
\end{align}
Since $\textsc{cmi}^{ij}$ measures how using $s^j_t$ could additionally contribute to predicting $s^i_{t+1}$ given the other state and action variables, $x^{\neg j}_t$, $\colorphi$ is a \textcolor{green}{\textbf{conditioned}} model trained to capture the additional information about $s^i_{t+1}$ in $s^j_t$ that is not present in $x^{\neg j}_t$. However, training $\phi$ and estimating $\textsc{cmi}^{ij}$ with Eq. (\ref{eq:cmi_implicit}) both require negative samples from $\textcolor{blue}{p(s^i_{t+1} | x^{\neg j}_t)}$, which are not readily accessible, as the data can only be collected from the full state transition distribution $\mathcal{T}(s_{t+1}|x_t)$.

\paragraph{Observation 2} To tackle this issue, CBM uses the importance sampling approximation proposed by \citet{sordoni2021decomposed} to compute $p(s^i_{t+1} | x^{\neg j}_t)$ with samples from the marginal distribution $p(s^i_{t+1})$. Thus, $\textsc{cmi}^{ij}$ may be approximated as,
\begin{align}
&\mathop{\mathbb{E}}\left[\log\frac{(N+1) e^{\phi(s^i_{t+1}; s_t^j | x^{\neg j}_t)}}{e^{\phi(s^i_{t+1}; s_t^j | x^{\neg j}_t)} + N\sum_{n=1}^N \textcolor{orange}{w_n} e^{\phi( \tilde{s}^{i, n}_{t+1}; s_t^j | x^{\neg j}_t)}}\right],\nonumber \\
&\textcolor{orange}{w_n}=
\frac{e^{\colorpsi(\tilde{s}^{i, n}_{t+1};x^{\neg j}_t)}}{\sum_{k=1}^N e^{\colorpsi(\tilde{s}^{i, k}_{t+1};x^{\neg j}_t)}}\approx \frac{p(s^i_{t+1} | x^{\neg j}_t)}{p(s^i_{t+1})}.
\label{eq:cmi_demi}
\end{align}
where $\tilde{s}^{i, n}_{t+1} \sim \textcolor{blue}{p(s^i_{t+1})}$, and $\colorpsi$ is trained to extract information about $s^i_{t+1}$ from $x^{\neg j}_t$ by minimizing $\mathcal{L}_\text{NCE}(\psi(s^i_{t+1}; x^{\neg j}_t))$ and is used to compute importance weights $w_n$ in a self-normalized manner.

After learning $\textcolor{orange}{\psi^*}$, one only needs $\colorphi$ to estimate CMI with Eq. (\ref{eq:cmi_demi}).
To this end, \citet{sordoni2021decomposed} train $\colorphi$ by minimizing $\mathcal{L}_\text{NCE}(\colorphi(s^i_{t+1}; s_t^j | x^{\neg j}_t) + \textcolor{orange}{\psi^*}(s^i_{t+1}; x^{\neg j}_t))$ while keeping $\psi^*$ frozen, so $\phi$ learns to capture the \textit{additional} information about $s^i_{t+1}$ from $s^j_t$ that is not present in $x^{\neg j}_t$ (i.e., absent in $\textcolor{orange}{\psi^*}$). See pseudo-code in the Appendix for details.

\begin{figure}[t]
\centering
\includegraphics[width=0.8\columnwidth]{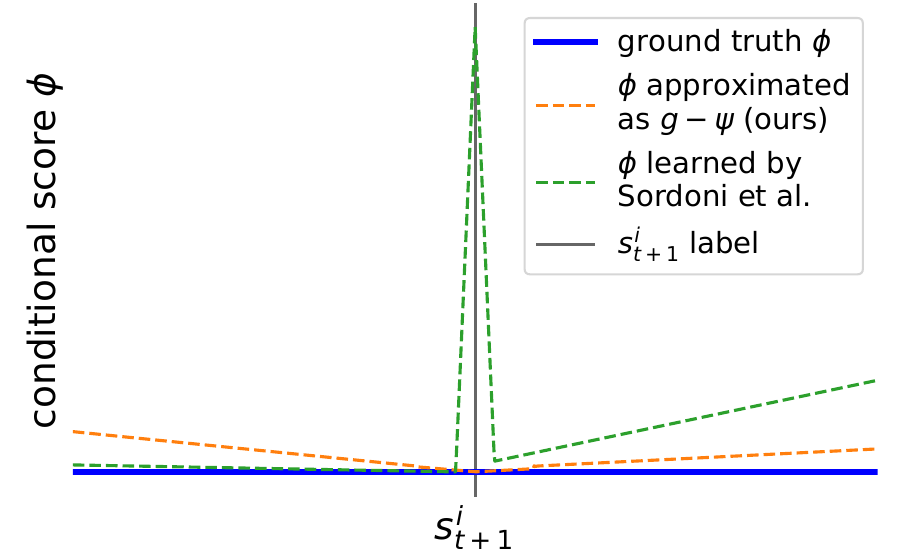}
\vspace{-10pt}
\includegraphics[width=0.8\columnwidth]{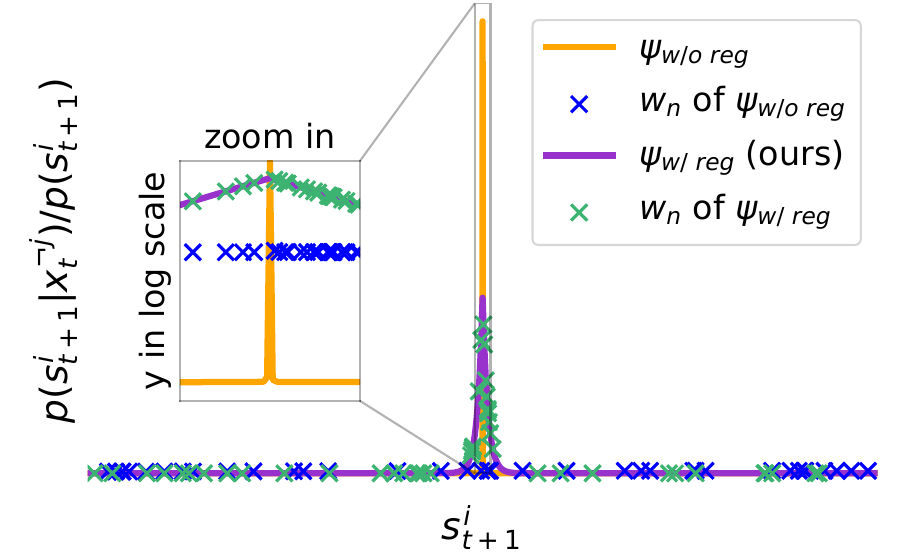}
\caption{\small Two sources of inaccurate CMI estimation: (\textbf{top}) Overfitted conditional model: learned $\colorphi$ overestimate conditional information while the $g - \colorpsi$ approximation is closer to the ground truth (0 for all $s_{t+1}^i$ values), especially when close to the ground truth label $s_{t+1}^i=0$. (\textbf{bottom}) Inaccurate Importance Sampling: without regularization, the likelihood ratio of $p(s^i_{t+1} | x^{\neg j}_t) / p(s^i_{t+1})$ computed by $\colorpsi$ can be peaked at the label $s_{t+1}^i=0$ and is therefore challenging to approximate by self-normalized importance sampling. In comparison, regularized $\psi$ has a flatter likelihood ratio landscape and is easier to approximate with samples.}
\label{fig:cmi_estimation_issues}
\vspace{-10pt}
\end{figure} 

\subsubsection{Inaccurate CMI Estimation and Solutions}
\label{subsubsec:cmi_overestimation}

In practice, the method proposed by \citet{sordoni2021decomposed} often yields inaccurate CMI estimations and thus leads to incorrect causal dependencies. We discovered two reasons for the inaccuracy and proposed corresponding solutions.

\paragraph{Reason 1 -- Overfitted Conditional Models}
In theory, when $\colorphi$ is trained as described above, it should condition on $\colorpsi$ and capture the \textit{additional} information only. However, in practice, such trained $\phi$ still uses $x^{\neg j}_t$ to predict $s^i_{t+1}$ directly rather than conditioning on it to estimate the additional contribution of $s^j_t$. 
Fig. \ref{fig:cmi_estimation_issues} top shows an example where knowing $s^j_t$ does not provide any additional information about $s^i_{t+1}$. 
In such a case, the ground truth conditional model should output the same score for all $s^i_{t+1}$ values. In contrast, the scores output by $\phi$ are still peaked at the label of $s^i_{t+1}$, and using such an overfitted model in Eq. (\ref{eq:cmi_demi}) would overestimate CMI.

To solve this issue, rather than using a learned $\phi$, we use the approximation $\colorphi = g - \colorpsi$. The motivation is as follows: as $g$ is trained to use $x_t$ to estimate the score of $s^i_{t+1}$ and $\psi$ estimates with all variables except for $s^j_t$, the difference of their estimated scores should reflect the additional information from $s^j_t$. In practice, as shown in Fig. \ref{fig:cmi_estimation_issues} top, the conditional score estimated by this approximation is closer to the ground truth than the score of $\colorphi$ learned by \citet{sordoni2021decomposed}, especially in the neighbor of the ground truth label where the accuracy of conditional scores significantly influences CMI.

\paragraph{Reason 2 -- Inaccurate Importance Sampling}
Meanwhile, when $s^i_{t+1}$ has an almost deterministic transition (which is common in many environments, e.g., objects will not move unless manipulated by the robot), the importance sampling approximation in Eq. (\ref{eq:cmi_demi}) could be inaccurate.

In detail, as shown in Fig. \ref{fig:cmi_estimation_issues} bottom, for such transitions, the score estimated by $\colorpsi$ tends to have extremely sharp maxima --- it is high only when $s^i_{t+1}$ is very close to the ground truth labels. 
As a result, even with many negative samples from $p(s^i_{t+1})$, it is likely that none of them are similar enough to samples from $p(s^i_{t+1} | x^{\neg j}_t)$. Then, since the importance weight $w_n$ in Eq. (\ref{eq:cmi_demi}) is self-normalized among all negative samples, samples that are not from $p(s^i_{t+1} | x^{\neg j}_t)$ still have large weights (rather than near-zero weights), thus leading to inaccurate CMI estimation. 

To mitigate this issue, when training $g$ and $\colorpsi$ for the dynamics of $\mathcal{S}^i_{t+1}$, beyond the InfoNCE loss, we regularize them to have flatter score landscapes with L2 penalty on their computed scores and the partial derivative of scores as follows, 
\begin{align}
\mathcal{L}_\text{dyn} = &\mathcal{L}_\text{reg}(g) + \mathcal{L}_\text{reg}(\colorpsi) \text{ where for generic $f$, }\label{eq:dyn_loss}\\
\mathcal{L}_\text{reg}(f) = &\mathcal{L}_\text{NCE}(f(s^i_{t+1};\cdot)) + \nonumber\\
&\sum_{\tilde{s}^i_{t+1}} \left(\lambda_1 \norm{f(\tilde{s}^i_{t+1}; \cdot)}^2 + \lambda_2 \norm{\frac{\partial f(\tilde{s}^i_{t+1}; \cdot)}{\partial \tilde{s}^i_{t+1}}}^2 \right)\nonumber.
\end{align}
The regularization is applied to both the label and all negative samples (i.e., $\tilde{s}^i_{t+1} \in \{s^i_{t+1}, \tilde{s}^{i, n}_{t+1}\}$), and $\lambda_1$, $\lambda_2$ are the weights of the regularization terms. 
As shown in Fig. \ref{fig:cmi_estimation_issues} bottom, with regularization, $\colorpsi$ is flatter. As a result, with the same number of samples, the importance weights $w_n$ approximate the likelihood ratio computed by $\psi$ much better, compared to approximating sharp $\psi$ without regularization.

To make the dynamics model scalable, we use the same masking technique as in Sec. \ref{subsec:causal_reward_learning} to combine $g$ and $\psi$ for each $j$ into one network. We use $\theta_\text{dyn}$ to denote the parameters of $d_\mathcal{S}$ such networks, each modeling the dynamics of state variable $\mathcal{S}^i_{t+1}$.

\subsection{CBM for Task Learning}
\label{subsec:task_learning}

\vspace{-10pt}
\begin{algorithm}[H]
  \caption{Causal Bisimulation Modeling (CBM)}
  \label{alg:mbrl}
  \begin{algorithmic}[1]
    \STATE Initialize the dynamics model $\theta_\text{dyn}$. 
    \STATE (Optional) Pretrain $\theta_\text{dyn}$ (Eq. \ref{eq:dyn_loss}) with offline data.
    \FOR{K tasks}
        \STATE Initialize the reward model 
        $\theta_\text{rew}$ and the policy $\pi$.
        \FOR{T training steps}
            \STATE Collect $(s_t, a_t, r_t, s_{t+1})$ with $a_t \sim \pi$.
            \STATE Update $\theta_\text{dyn}$ (Eq. \ref{eq:dyn_loss}, optional) and $\theta_\text{rew}$ (Sec. \ref{subsec:causal_reward_learning}).
            \STATE Evaluate dynamical and reward dependencies (Eq. \ref{eq:cmi_demi}); Update the state abstraction for $\pi$ (Fig. \ref{fig:causal_illustration} (b)).
            \STATE Update $\pi$ SAC losses.
        \ENDFOR
    \ENDFOR
  \end{algorithmic}
\end{algorithm}
\vspace{-10pt}

As shown in Alg. \ref{alg:mbrl}, for tasks with the same dynamics, CBM's dynamics model is shared across tasks. The dynamics model can either learn from offline data (line 2), or from transitions collected during task learning (line 7), or both. 

When solving each task, CBM interweaves reward learning (line 7) with policy learning. The policy is trained via Soft Actor Critic (SAC, \citet{Haarnoja2018SoftAC}), an off-policy reinforcement learning algorithm. 
The reward model is combined with the pre-trained dynamics to generate the task-specific abstraction (line 8). 
CBM applies the state abstraction to the policy $\pi$ as a binary mask that zeros out ignored variables. 
During task learning, as the policy explores and learns, we expect it to gradually expose causal relationships that are necessary to solve the task, and, in return, the updated state abstractions reduce the learning space of the policy, making its learning sample efficient.

%% file: contents/06_experiments.tex
\section{Experiments}

\begin{table*}[htbp]
\centering
\small
\begin{tabular}{lcccccccc}
\toprule
                & \multicolumn{3}{c}{block}  & & \multicolumn{2}{c}{tool-use}                        
                \\\cline{2-4} \cline{6-7}
                & causal graph         & pick         & stack & & causal graph    & series      
                \\ \midrule
    explicit        & 87.5 $\pm$ 0.1   & 53.2 $\pm$ 4.6  & 59.6 $\pm$ 4.6    & & 82.6 $\pm$ 0.2  & 80.0 $\pm$ 1.5            
    \\
    implicit (ours) & \textbf{90.5} $\pm$ 0.4  & \textbf{95.7} $\pm$ 6.0   & \textbf{95.7} $\pm$ 6.0   & & \textbf{85.5} $\pm$ 0.1 & \textbf{98.8} $\pm$ 1.3   
    \\ 
    \bottomrule
\end{tabular}
\vspace{-5pt}
\caption{Mean $\pm$ std. error of accuracy ($\uparrow$) for learned dynamics causal graphs and task abstractions.}
\vspace{-10pt}
\label{tbl:causal_graph}
\end{table*}
\label{sec:experiments}

We examine the following hypotheses. 
First, implicit models recover dynamical dependencies more accurately than explicit models (Sec. \ref{exp:dynamics}).
Second, compared to CDL's task-independent state abstraction and prior task-specific abstraction works, CBM learns a more concise abstraction and improves sample efficiency and generalization of task learning over baselines (Sec. \ref{exp:task}).

\paragraph{Environments}
To test CBM, we use two manipulation environments implemented with Robosuite \citep{robosuite2020}, shown in Fig. \ref{fig:task_policy_learning} left, and two tasks from the DeepMind Control Suite (DMC, \citet{tunyasuvunakool2020}). In the block environment (b), there are multiple movable and unmovable blocks. The tasks in this environment include Pick and Stack.
In the tool-use environment, we consider a challenging long-horizon task Series: the agent needs to use an L-shaped tool to move a faraway block within reach, pick it up, and place it within the box. 
In the DMC, we consider the Cheetah and Walker tasks, two high-dimensional continuous control tasks. 
In all environments, as controllable distractors ($cd$), 20 variables whose values are random projections of the action (i.e., $W^T a_t $ where $W \subseteq \mathbb{R}^{d_\mathcal{A}}$ are randomly sampled) are added to the state space. We also add 20 uncontrollable distractors ($ud$) whose values are uniformly sampled from $[-1, 1]$. The distractors have no interaction with other state variables. 

\paragraph{Baselines}
For the dynamics learning experiments, we compare the implicit dynamics model against the explicit model.  All methods are trained and evaluated with 3 random seeds.
For the state abstraction and task learning experiments, we compare CBM against the closely related methods of CDL, which uses a task-independent abstraction; TIA \citep{fu2021learning} and Denoised MDP \citep{pmlr-v162-wang22c}, which both learn task-specific state abstractions. To contextualize these methods, we also compare against an Oracle that learns with the ground-truth minimal abstractions, and reinforcement learning over the Full state space (no state abstraction). All methods are trained and evaluated with 5 random seeds.
Further details are in the Appendix.

\paragraph{State Abstractions for Task Learning}
To fairly compare the effects of the various state abstractions for task learning, all methods use \textbf{implicit} dynamics models, including CDL, which originally used the inferior explicit models. All methods learn tasks via Soft Actor Critic \citep{Haarnoja2018SoftAC}. The dynamics model is pretrained in Pick and Stack tasks, and it is learned jointly with the policy \textit{from scratch} in all other tasks. Further details are in the Appendix.

\subsection{Dynamics and Causal Graph Learning}
\label{exp:dynamics}

This section compares implicit and explicit dynamics models, learned from the same offline data, in terms of how accurately they recover dynamics causal graphs and task abstractions. Experiments are conducted on the Robosuite environments\footnote{DMC tasks are not suitable for these experiments, as the ground truth causal graphs are not clear.}, and results are shown in Table \ref{tbl:causal_graph}.
For causal graphs, we compare the learned dynamics dependencies with the ground truth and measure the \textit{causal graph accuracy }as the \# correctly learned edges / total edges in the graph. The accuracy of implicit models is 3\% higher than that of explicit ones, which corresponds to causal relationships between 67 pairs of state variables in the block environment, and 48 pairs in tool-use. 
As a result, when being used to derive a \textit{task-specific} abstraction (we use the ground truth reward dependencies in this experiment only to avoid the interference from reward learning), implicit models derive more accurate state abstractions ($\ge 20\%$ difference in accuracy on all tasks) than explicit models. Note that the \textit{state abstraction accuracy} is measured as the \# of correctly classified state variables / total state variables. Implicit models also learn more generalizable dynamics (see Appendix).

\subsection{Task Learning with State Abstractions}
\label{exp:task}

This section compares policy learning with different state abstractions on various tasks in the DMC, the block environment (b), and the tool-use environment (t).

\paragraph{State Abstraction Accuracy}

\begin{figure}[t]
  \centering
  \includegraphics[width=0.9\columnwidth]{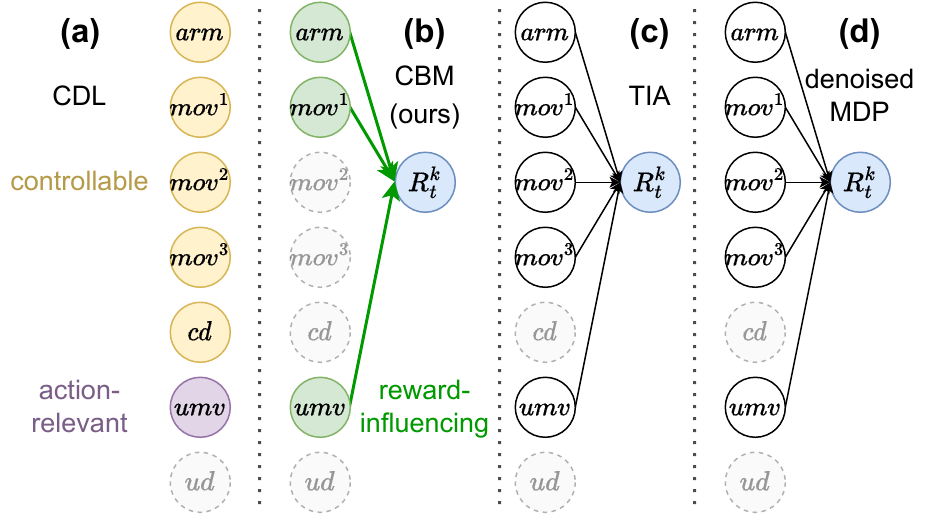}
  \vspace{-5pt}
  \caption{\small Object-level state abstractions for Stack learned by each method; semitransparent variables are excluded. \textbf{(a)} Though $mov^2$, $mov^3$, and controllable distractos $cd$ are unnecessary, CDL still keeps them as they are \textcolor{cdl_yellow}{controllable}. \textbf{(b)} Compared to CDL, CBM (ours) successfully learns the minimal abstraction by further reasoning which state variables \textcolor{cbm_green}{influence} the reward.  \textbf{(c)} TIA and \textbf{(d)} Denoised MDP fail to learn meaningful abstractions when their assumptions on the dynamics do not hold.}
  \label{fig:abstraction}
  \vspace{-15pt}
\end{figure}

Fig. \ref{fig:abstraction} shows the abstraction learned by each method for the Stack task. For simplicity, abstractions are shown on the object level; state variable-level abstractions are in the Appendix. 
Among all methods, only CBM learns minimal abstraction. 
CDL keeps all controllable variables and thus uses a non-minimal abstraction. 
Meanwhile, TIA and Denoised MDP assume that state variables can be segregated into several dynamically independent components, and their abstractions keep only the task-relevant component.
However, though $mov^2$ and $mov^3$ are task-irrelevant, their dynamics still depend on the task-relevant part (the end-effector and gripper). As a result, when their assumptions do not hold, TIA and Denoised MDP learn that (almost) all variables belong to the same component. Then, depending on their definitions of task relevance, all such variables are either included (TIA) or ignored (Denoised MDP) by the abstractions.

\paragraph{CBM is Sample-Efficient}

\begin{figure*}[ht]
\centering
\begin{subfigure}[b]{0.48\columnwidth}
\centering
\includegraphics[width=0.78\linewidth]{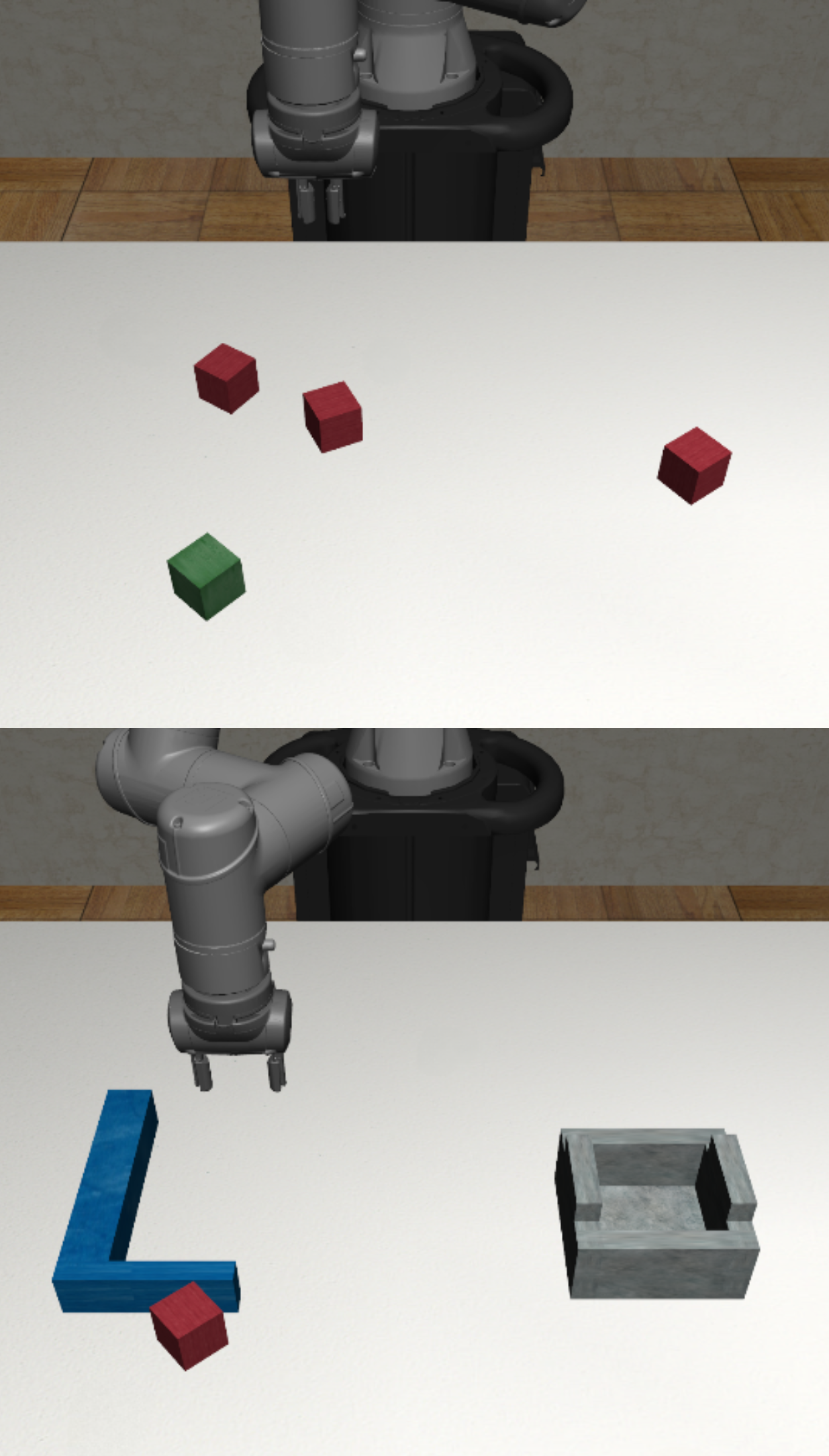}
\end{subfigure}
\begin{subfigure}[b]{1.5\columnwidth}
\centering
\includegraphics[width=0.8\linewidth]{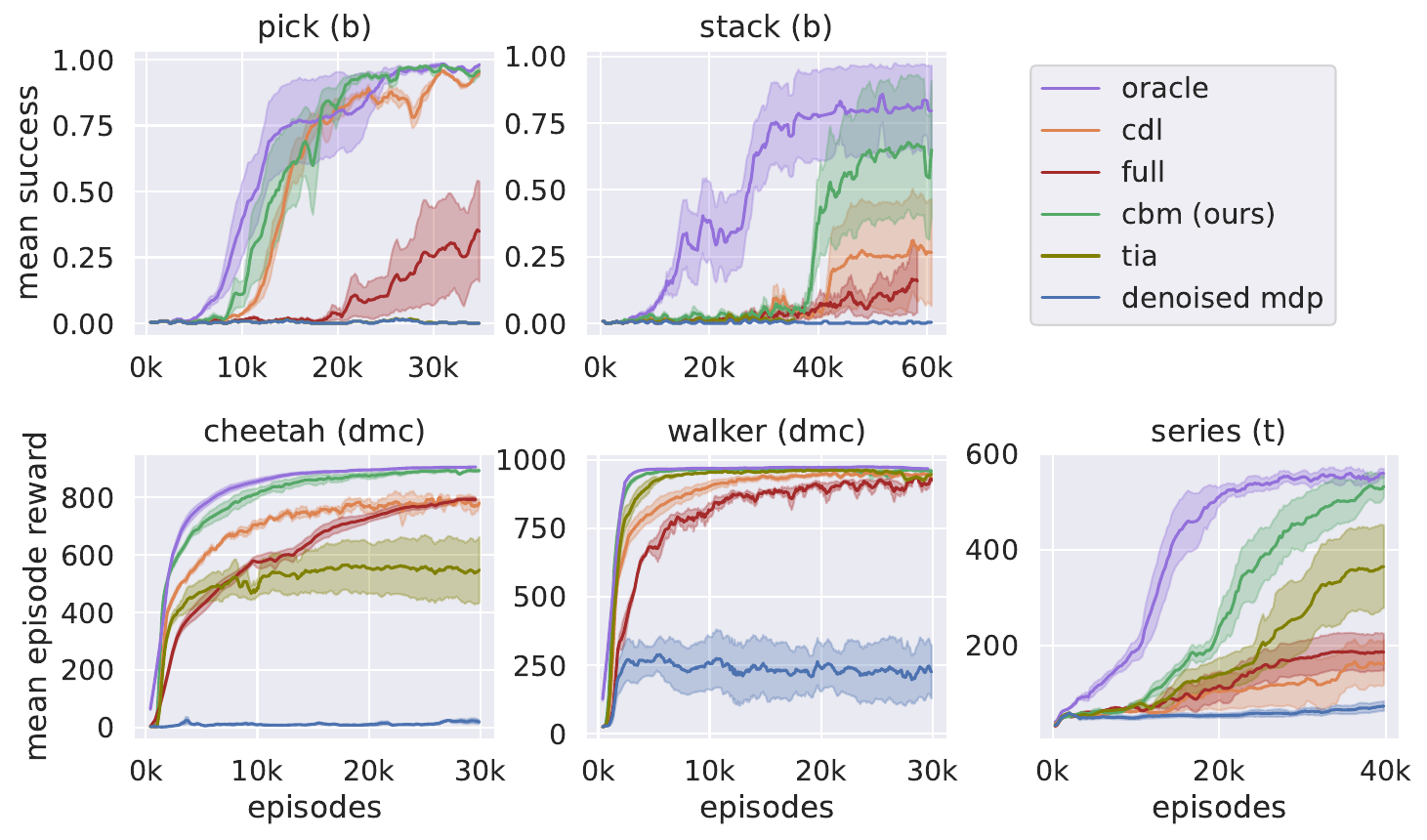}
\end{subfigure}
\vspace{-5pt}
\caption{\small 
(\textbf{left}) top: Block environment with three \textcolor{red}{\textit{movable}} blocks ($mov^{1 \sim 3}$), and an \textcolor{green}{\textit{unmovable}} ($unm$) block fixed on the table.
bottom: Tool-Use environment with the \textcolor{red}{block}, the \textcolor{blue}{L-shaped tool}, and the \textcolor{grey}{box}.
(\textbf{right}) Learning curves of CBM (ours) compared to baseline methods 
and RL with the Oracle abstraction in five tasks. Each learning curve is generated from independent runs using 5 different random seeds, with mean and std. error computed across 50 test episodes per point on the learning curve. CBM is among the most sample-efficient methods, even approaching the efficiency of the Oracle on Pick, Cheetah, and Walker.}
\label{fig:task_policy_learning}
\vspace{-10pt}
\end{figure*}

The performance metric for Pick and Stack tasks is the mean success rate at accomplishing the task, and the metric for Series and DMC tasks is the mean episode reward, evaluated over 50 test episodes and plotted with respect to the number of episodes.\footnote{As each episode is a fixed number of steps, episodes have a linear relationship with training steps.} 
The learning curves for all tasks are shown in Fig. \ref{fig:task_policy_learning} right. Recall that Oracle learns with the ground-truth state abstraction, whereas Full learns with no state abstraction. For all tasks, Oracle learns the fastest, demonstrating the possible gain in sample efficiency with an ideal state abstraction.

Overall, we find that CBM matches or improves in sample efficiency over the CDL, TIA, and Denoised MDP baselines in all settings. In all tasks, CBM is among the closest to the Oracle in sample efficiency, showing the benefit of the learned, near-minimal state abstraction.
We observe that the higher the difficulty level of the task, the greater the benefit of learning a small task abstraction. 
For instance, Walker is a simple task, where almost all methods converge rapidly to an episode reward of 1000 by 20k episodes. The only exception is Denoised MDP, where the learned abstraction masks out some key variables among robot joint angles and angular velocities. On the other hand, the Series task is much more challenging, requiring a sequence of successful behaviors (reach tool, use tool to move block closer, pick and place block). Learning without any abstraction (Full) only learns to grasp the tool and achieves an episode reward of 200 over 40k training episodes, while CBM learns with much greater sample efficiency. We observe a similarly large gap between CBM and other methods on Stack (B), another complex manipulation task. An ablation of CBM with explicit dynamics models can be found in the Appendix; we find that the ablation has much worse sample efficiency than CBM, demonstrating the benefit of using implicit dynamics models.

\begin{figure}[htbp]
  \centering
\includegraphics[width=1.0\columnwidth]{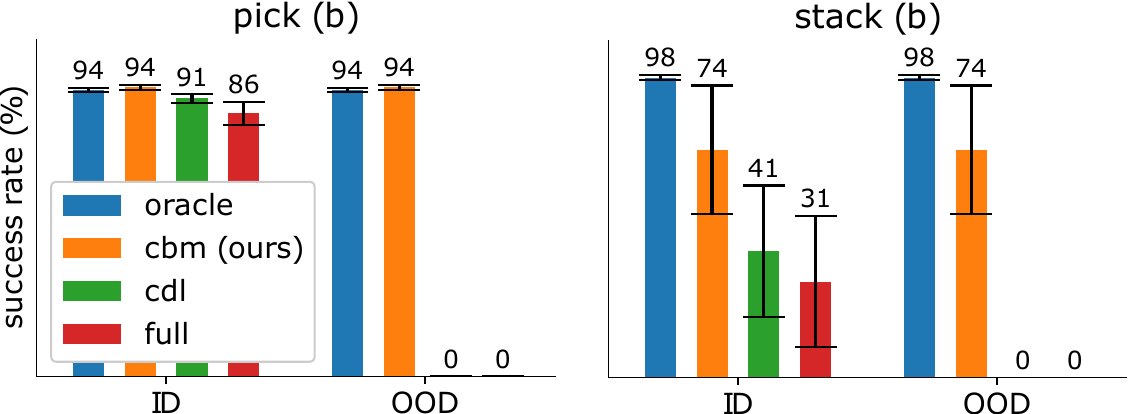}
  \vspace{-20pt}
  \caption{\small Performance of policies with different state abstractions on both ID and OOD states, in terms of mean and std. error of success rates ($\uparrow$) in the block environment.}
  \label{fig:generalization}
  \vspace{-10pt}
\end{figure}

\paragraph{CBM is Generalizable}
As shown in Fig. \ref{fig:generalization}, we further measure each method's generalizability to unseen states on Pick and Stack (TIA and Denoised MDP fail to learn the tasks and thus are not evaluated). In addition to in-distribution (ID) states, we also evaluate the learned policies on out-of-distribution (OOD) states where the values of all task-irrelevant state variables are set to noise sampled from $\mathcal{N}(0, 1)$. 
For both tasks, only Oracle and CBM keep similar performance across ID and OOD states, as their minimal task-specific abstractions eliminate the influence of OOD variables. In contrast, though CDL can be robust against variables ignored by its abstraction, it still fails to generalize when redundant variables in its non-minimal abstractions have unseen values.

%% file: contents/07_conclusions.tex
\section{Conclusion}
\label{sec:conclusions}

This paper studies how to generate task-specific minimal state abstractions for task learning. 
It introduces Causal Bisimulation Modeling (CBM), an algorithm that (1) learns a minimal state abstraction via causal reward learning, and (2) learns an implicit causal dynamics model.
The experiments demonstrate that CBM learns more accurate and concise state abstractions, which lead to improved sample efficiency on downstream tasks compared to related methods. Further, the implicit dynamics model introduced by CBM improves over explicit dynamics models in terms of prediction error and causal graph accuracy. Promising directions for future work include relaxing the assumption of having a pre-defined factored state space to extend CBM to high-dimensional state spaces, such as images. 

%% file: contents/08_appendix.tex
\onecolumn
\textbf{\Large Appendix}

\section{Pseudo-code for Sec \ref{subsec:causal_learning}}

In this section, we provide pseudo-code for how \citet{sordoni2021decomposed} and CBM learn the conditional implicit model $\colorphi$, respectively, in Alg \ref{alg:demi_phi_learning} and Alg \ref{alg:cbm_phi_learning}. For easier reference, first, we reproduce key equations and notations as follows:

\paragraph{InfoNCE Loss} 
For a generic model $f: \mathcal{Y} \times \mathcal{Z} \rightarrow \mathbb{R}$ that tries to extract information about a generic variable $y \in \mathcal{Y}$ from another variable $z \in \mathcal{Z}$. The corresponding InfoNCE loss of $f$ is
\begin{align}
    \mathcal{L}_\text{NCE}(f(y;z)) &= -\log \frac{e^{f(y; z)}}{e^{f(y; z)} + \sum_{n=1}^N e^{f(\tilde{y}^{n}; z)}}, \text{ where negative samples are drawn from } \tilde{y}^{n} \sim \mathcal{Y}. \nonumber
\end{align}

\paragraph{Involved Implicit Models} 
In Sec \ref{subsec:causal_learning}, when \citet{sordoni2021decomposed} and CBM estimate conditional mutual information $\textsc{cmi}^{ij}$ between $s^j_t$ and $s^i_{t+1}$ respectively, the following three implicit models are involved:
\begin{itemize}
    \item $g(s^i_{t+1};x_t): \mathcal{S}^i \times \mathcal{X} \rightarrow \mathbb{R}$, extracting information about $s^i_{t+1}$ from $x_t = (s_t, a_t)$, where $\mathcal{X} = \mathcal{S} \times \mathcal{A}$.
    \item $\colorpsi(s^i_{t+1};x^{\neg j}_t): \mathcal{S}^i \times \mathcal{X}^{\neg j}\rightarrow \mathbb{R}$, extracting information about $s^i_{t+1}$ from $x^{\neg j}_t = (s^1_t, \dots, s^{j-1}_t, s^{j+1}_t, \dots, s^{d_\mathcal{S}}_t, a_t)$, where $\mathcal{X}^{\neg j} = \mathcal{S}^1 \times \cdots \times \mathcal{S}^{j-1}_t \times \mathcal{S}^{j+1}_t \times \cdots \times \mathcal{S}^{d_\mathcal{S}}_t \times \mathcal{A}$.
    \item $\colorphi(s^i_{t+1}; s_t^j | x^{\neg j}_t): \mathcal{S}^i \times \mathcal{S}^j \times \mathcal{X}^{\neg j} \rightarrow \mathbb{R}$, extracting the \textcolor{green}{\textbf{additional}} information about $s^i_{t+1}$ in $s^j_t$ that is not present in $x^{\neg j}_t$.
\end{itemize}

\vspace{-5pt}
\begin{algorithm}[h!]
  \caption{$\colorphi$ learned by \citet{sordoni2021decomposed}}
  \label{alg:demi_phi_learning}
  \begin{algorithmic}[1]
    \STATE Initialize the $\colorpsi$ and $\colorphi$.
    \REPEAT
        \STATE Sample $s^i_{t+1}$ and $x^{\neg j}_t$ from data; Sample $\{\tilde{s}^{i, n}_{t+1}\}_{n=1}^N$ from $\mathcal{S}^i$.
        \STATE Optimize $\colorpsi$ with $\mathcal{L}_\text{NCE}(\colorpsi(s^i_{t+1}; x^{\neg j}_t))$.
    \UNTIL{$\colorpsi$ converges to $\textcolor{orange}{\psi^*}$}
    \REPEAT
        \STATE Sample $s^i_{t+1}$, $s^j_t$, $x^{\neg j}_t$ from data; Sample $\{\tilde{s}^{i, n}_{t+1}\}_{n=1}^N$ from $\mathcal{S}^i$.
        \STATE Optimize $\colorphi$ with the following loss while \textbf{keeping} $\textcolor{orange}{\psi^*}$ \textbf{frozen}
        \begin{equation}
            \mathcal{L}_\text{NCE}(\colorphi + \textcolor{orange}{\psi^*}) = -\log \frac{e^{\colorphi(s^i_{t+1}; s_t^j | x^{\neg j}_t) + \textcolor{orange}{\psi^*}(s^i_{t+1}; x^{\neg j}_t)}}{e^{\colorphi(s^i_{t+1}; s_t^j | x^{\neg j}_t) + \textcolor{orange}{\psi^*}(s^i_{t+1}; x^{\neg j}_t)} + \sum_{n=1}^N e^{\colorphi(\tilde{s}^{i, n}_{t+1}; s_t^j | x^{\neg j}_t) + \textcolor{orange}{\psi^*}(\tilde{s}^{i, n}_{t+1}; x^{\neg j}_t)}}.
        \end{equation}
    \UNTIL{$\colorphi$ converges}
  \end{algorithmic}
\end{algorithm}

\vspace{-15pt}
\begin{algorithm}[h!]
  \caption{$\colorphi$ learned by CBM}
  \label{alg:cbm_phi_learning}
  \begin{algorithmic}[1]
    \STATE Initialize the $g$ and $\colorpsi$.
    \REPEAT
        \STATE Sample $s^i_{t+1}$ and $x_t$ from data; Sample $\{\tilde{s}^{i, n}_{t+1}\}_{n=1}^N$ from $\mathcal{S}^i$.
        \STATE Optimize $g$ with
        \begin{equation}
        \mathcal{L}_\textbf{dyn}(g(s^i_{t+1}; x_t)) = \mathcal{L}_\text{NCE}(g) + 
        \sum_{\tilde{s}^i_{t+1} \in \{s^i_{t+1}\} \cup \{\tilde{s}^{i, n}_{t+1}\}_{n=1}^N} \left(\lambda_1 \norm{g(\tilde{s}^i_{t+1}; x_t)}^2 + \lambda_2 \norm{\frac{\partial g(\tilde{s}^i_{t+1}; x_t)}{\partial \tilde{s}^i_{t+1}}}^2 \right). \nonumber
        \end{equation}
    \UNTIL{$g$ converges}
    \REPEAT
        \STATE Sample $s^i_{t+1}$ and $x^{\neg j}_t$ from data; Sample $\{\tilde{s}^{i, n}_{t+1}\}_{n=1}^N$ from $\mathcal{S}^i$.
        \STATE Optimize $\colorpsi$ with 
        \begin{equation}
        \mathcal{L}_\textbf{dyn}(\colorpsi(s^i_{t+1}; x^{\neg j}_t)) = \mathcal{L}_\text{NCE}(\colorpsi) + 
        \sum_{\tilde{s}^i_{t+1} \in \{s^i_{t+1}\} \cup \{\tilde{s}^{i, n}_{t+1}\}_{n=1}^N} \left(\lambda_1 \norm{\colorpsi(\tilde{s}^i_{t+1}; x^{\neg j}_t)}^2 + \lambda_2 \norm{\frac{\partial \colorpsi(\tilde{s}^i_{t+1}; x^{\neg j}_t)}{\partial \tilde{s}^i_{t+1}}}^2 \right). \nonumber
        \end{equation}
    \UNTIL{$\colorpsi$ converges}
    \RETURN $\colorphi(s^i_{t+1}; s_t^j | x^{\neg j}_t) = g(s^i_{t+1}; x_t) - \colorpsi(s^i_{t+1}; x^{\neg j}_t)$
  \end{algorithmic}
\end{algorithm}

After learning $\colorphi$, CBM computes $\textsc{cmi}^{ij}$ as in Alg. \ref{alg:cmi_compute}, following \citet{sordoni2021decomposed}.

\begin{algorithm}[ht]
  \caption{$\textsc{cmi}^{ij}$ computation using learned $\colorpsi$ and $\colorphi$}
  \label{alg:cmi_compute}
  \begin{algorithmic}[1]
    \STATE Sample $\{\tilde{s}^{i, n}_{t+1}\}_{n=1}^N$ from $\mathcal{S}^i$.
    \FOR{$n \in N$}
        \STATE Compute the self-normalized importance weight for each negative sample $\tilde{s}^{i, n}_{t+1}$ as
        \begin{equation}
            \textcolor{orange}{w_n}=
\frac{e^{\colorpsi(\tilde{s}^{i, n}_{t+1};x^{\neg j}_t)}}{\sum_{m=1}^N e^{\colorpsi(\tilde{s}^{i, m}_{t+1};x^{\neg j}_t)}}\approx \frac{p(s^i_{t+1} | x^{\neg j}_t)}{p(s^i_{t+1})}. \nonumber
        \end{equation}
    \ENDFOR
    \STATE Compute the conditional mutual information as
    \begin{equation}
        \textsc{cmi}^{ij} = \mathop{\mathbb{E}}\left[\log\frac{(N+1) e^{\colorphi(s^i_{t+1}; s_t^j | x^{\neg j}_t)}}{e^{\colorphi(s^i_{t+1}; s_t^j | x^{\neg j}_t)} + N\sum_{n=1}^N \textcolor{orange}{w_n} e^{\colorphi( \tilde{s}^{i, n}_{t+1}; s_t^j | x^{\neg j}_t)}}\right]. \nonumber
    \end{equation}
  \end{algorithmic}
\end{algorithm}

\section{Additional Comparison with Related Works}

Regarding the assumptions of the method and the quality of learned abstractions, below we provide a table comparing CBM against ICP \citep{zhang2020invariant}, CDL, TIA, and denoised MDPs in the following aspects:
\begin{itemize}
\item whether the method can learn in a single environment (SE)
\item whether the method can learn minimal state abstractions (MSA)
\item whether the learned dynamics can be shared across multiple tasks in the same environment (GD)
\item whether the method can learn from high-dimensional image observations (IO), instead of assuming that the state space is factored
\end{itemize}

\begin{table}[htbp]
\centering
\vspace{-5pt}
\small
\begin{tabular}{lcccc}
\toprule
                        & SE            & MSA           & GD            & IO \\ \midrule
CBM (ours)              & \checkmark    & \checkmark    & \checkmark    & $\times$ \\
ICP                     & $\times$      & $\times$      & $\times$      & \checkmark \\
CDL                     & \checkmark    & $\times$      & \checkmark    & $\times$     \\
TIA and denoised MDP    & \checkmark    & $\times$      & $\times$      & \checkmark \\
\bottomrule
\end{tabular}
\vspace{-0pt}
\label{tbl:related_works}
\end{table}

Overall, as described above, our method can learn minimal state abstractions, does not require multiple environments for training, and can share the dynamics across different tasks, but at the cost of assuming a factored state space.

\section{Theorem \ref{theorem:bisimulation} Assumptions}
\label{app:theorem}

Our MDP also follows the three assumptions of Theorem 1 in Zhang et al., (2020a). Specifically, 
(1) For Assumption 1 that each observation corresponds to a unique state, we assume the observation space is the state space and is fully observable. 
(2) For Assumption 2 that each observation component (i.e., each state variable in our setting) at $t+1$ is independent given observation at $t$, we have the same assumption in Sec 3.1.
(3) For Assumption 3 about the difference between environments in the same family, we only focus on a single environment and thus there is no need for this assumption. 

\section{Dynamics Learning Implementation and Further Results}
\label{app:dynamics_results}

\subsection{Implicit Dynamics Modelling Details}
\label{app:implicit_architectures}
We compute the energy $E/\psi(s^i_{t+1}; M \odot [s_t, a_t])$ as $g(M \odot [s_t, a_t])^T h(s^i_{t+1})$, where both $g$ and $h$ are multilayer perceptrons with three hidden layers of 128 units and outputs a size 128 vector. 
During training, we use regularization coefficients $\lambda_1 = \lambda_2 = 10^{-6}$. $\lambda_1$ and $\lambda_2$ are decided using grid search among $\{10^{-3}, 10^{-4}, \cdots, 10^{-7}\}$, by trading off prediction accuracy and causal graph accuracy.
The \textsc{cmi} threshold used to infer causal relationships is $\epsilon = 0.02$. 
During inference, to predict the next step value of each state variable $\hat{s}^i_{t+1}$ as $\argmax_{s^i_{t+1}}f(s_t, a_t, s^i_{t+1})$, we uniformly sample 8,192 samples from $\mathcal{S}^i$ and select the one with the lowest energy. We test the effector of the sample number in Table. \ref{tab:dyn_pred_cost}.

The architecture and hyperparameters of the implicit dynamics model are listed in Table \ref{tab:implicit_hyperparams}. We tested networks with 64/128/256 neurons in each layer and chose 128 as it achieves the same prediction performance as 256 with a shorter computation time. Similarly, we choose the number of negative samples as 512 from $\{256, 512, 1024\}$. Other hyperparameters are not tuned. In our experiments, instead of predicting the energy based on $(M \odot [s_t, a_t])$ and $s^i_{t+1}$, we use $(M \odot [s_t, a_t])$ and $\Delta s^i_t$ where $\Delta s^i_t$ is the change of the variable at $t$. Then, for negative samples, we sample them from $[\Delta s^i_{\min}, \Delta s^i_{\max}]$.

\begin{table}[ht]
\centering
\caption{Hyperparameters for the implicit dynamics model.}
\vspace{-0pt}
\begin{small}
\begin{tabular}{cc}
\toprule
\textbf{Name}         & \textbf{Value} \\
\midrule
feature architecture          & [128, 128]       \\
energy architecture           & [128]            \\
activation functions          & ReLU             \\
number of training transitions          & 2M            \\
training step                           & 3M             \\
number of negative samples              & 512                  \\
learning rate                           & 3e-4                  \\
batch size                              & 32                    \\
prediction step during training, $H$    & 3                     \\
\bottomrule
\end{tabular}
\end{small}
\vspace{-0pt}
\label{tab:implicit_hyperparams}
\end{table}

\subsection{Dynamics Data Collection}

CBM proposes a novel method to evaluate the causal dependencies with implicit models, recovering more accurate dependencies than explicit models. However, in addition to the causal discovery method, the quality of the recovered causal relationship also depends on the collected data. In this section, we explain how we ensure the data is collected by diverse policies to enable correct inference of causal dependencies.

If the dynamics model is trained on offline data only, the data should be collected by a diverse set of policies to break the spurious correlations.
We may not have such offline data. For example, offline data are collected by a single policy only. In this case, we can augment the data with online data collected by the set $\{\pi_n^k\}_{k=1,\dots, K; t=1,\dots, T}$, where $k$ in the task index and $t$ is the training step. This set of policies will naturally be diverse, because:
\begin{itemize}
    \item Considering $k$, each $\pi_t^k$ learns from a different reward and behaves uniquely.
    \item Considering $t$, as $\pi_t^k$ explores and gets updated, its behavior also changes and thus the data is being collected by different policies.
    \item Though Alg. \ref{alg:mbrl} shows that we learn tasks sequentially for simplicity of presentation, in practice, we can learn all tasks simultaneously so that the dynamics model can use data collected by all policies for learning.
\end{itemize}

\subsection{Learned Dynamics Causal Graph}
\label{app:causal_graph_results}

\begin{figure}[hbtp]
  \centering
  \includegraphics[width=1.0\linewidth]{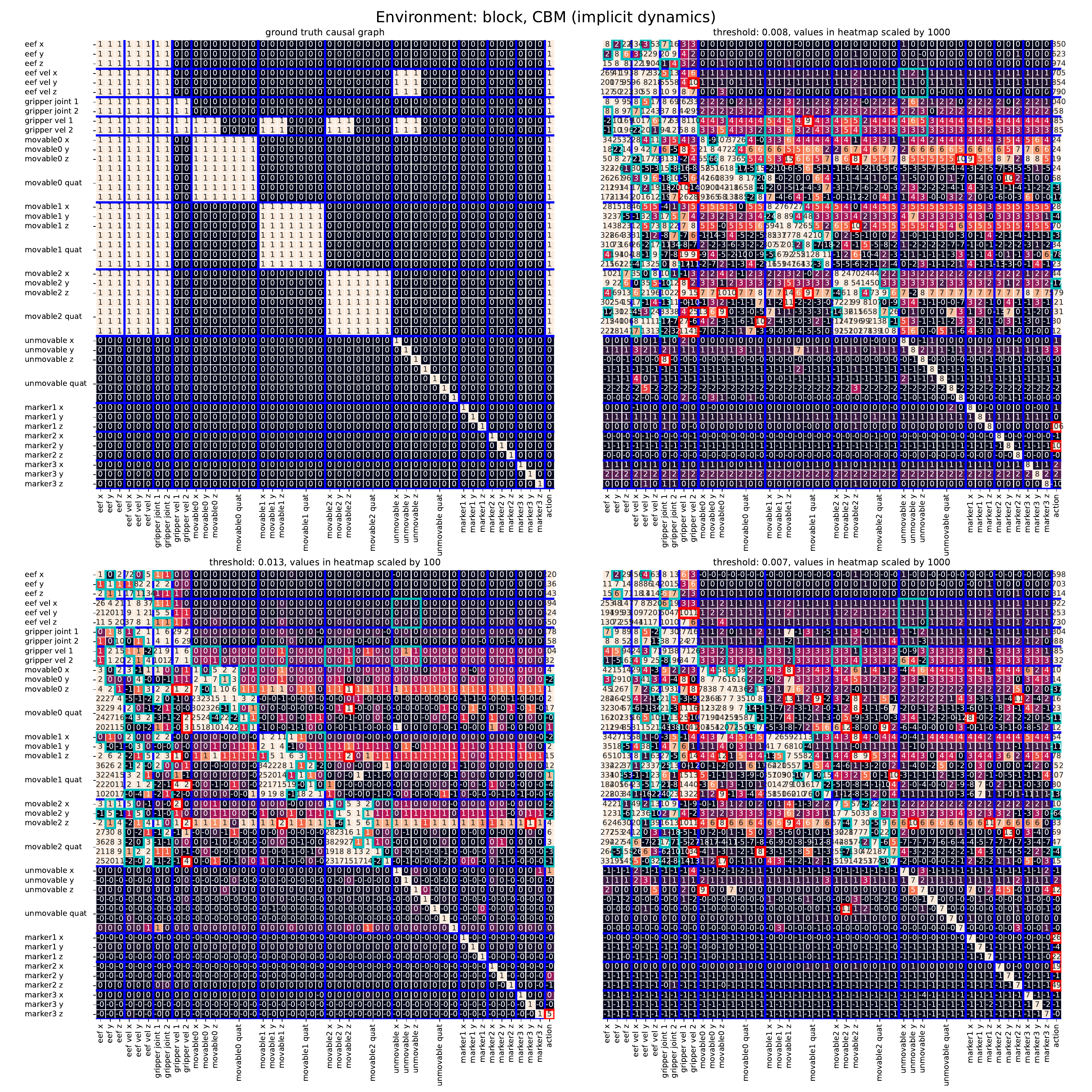}
  \caption{\small The ground truth dynamics causal graph and causal graphs learned by CBM in the block environment across 3 seeds.}
  \label{fig:causal_graph_block_implicit}
  \vspace{-0pt}
\end{figure}

\begin{figure}[hbtp]
  \centering
  \includegraphics[width=1.0\linewidth]{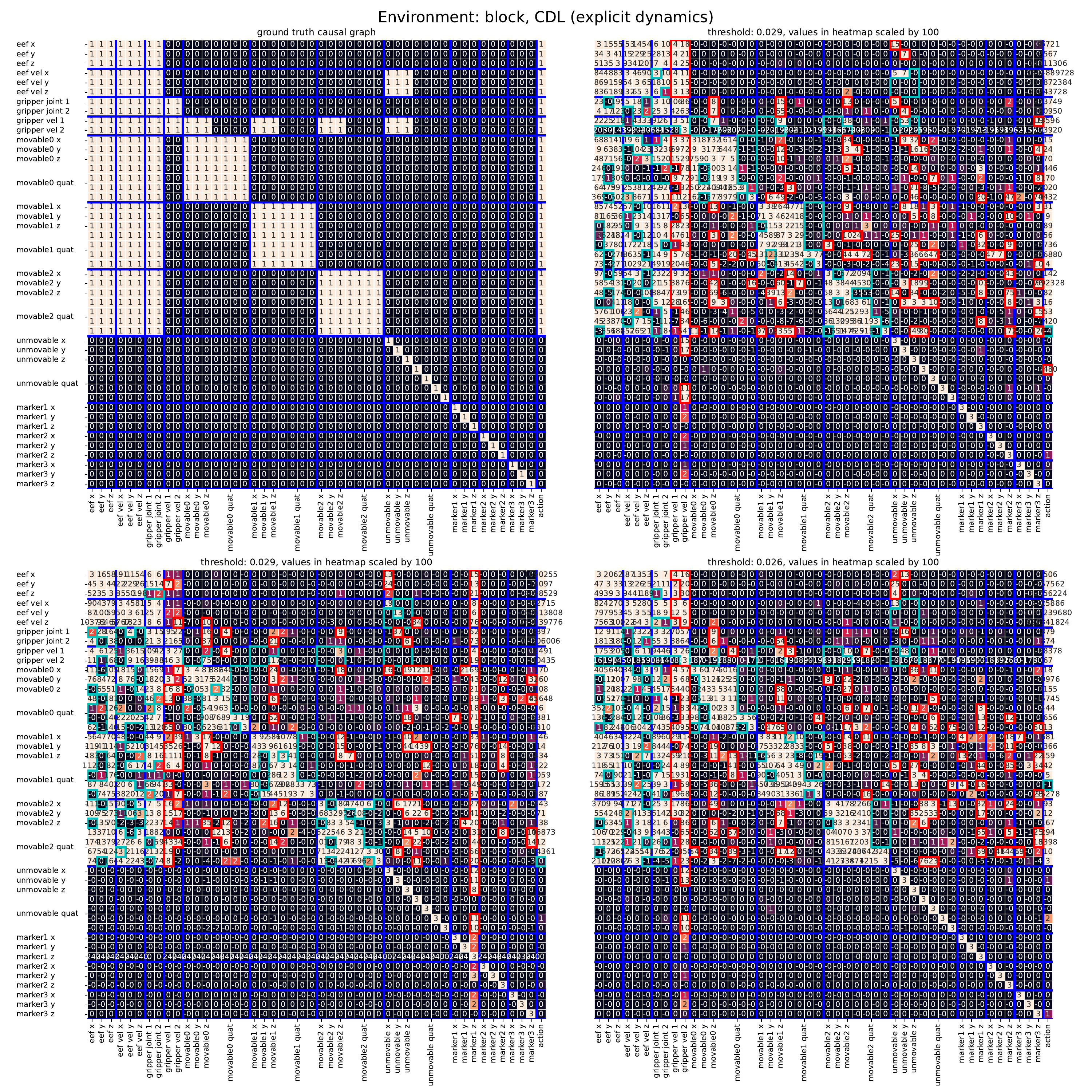}
  \caption{\small The ground truth dynamics causal graph and causal graphs learned by CDL in the block environment across 3 seeds.}
  \label{fig:causal_graph_block_explicit}
  \vspace{-0pt}
\end{figure}

In addition to the causal graph accuracy discussed in Table. \ref{tbl:causal_graph}. We also show the dynamics causal graphs learned by implicit and explicit dynamics for the block environment, in Fig. \ref{fig:causal_graph_block_implicit} and Fig. \ref{fig:causal_graph_block_explicit}, respectively. The strength of the causal dependencies are measured in \textsc{cmi} and numbered in each cell. Meanwhile, the missing dependencies are marked in red while spurious ones are marked in green. We notice that implicit dynamics models tend to miss dependencies that are necessary but happen infrequently, while explicit models tend to depend on more spurious correlations and thus generalize badly in out-of-distribution states.

\subsection{Dynamics Prediction}
\label{app:prediction_results}

\begin{figure}
\centering
\includegraphics[width=\textwidth]{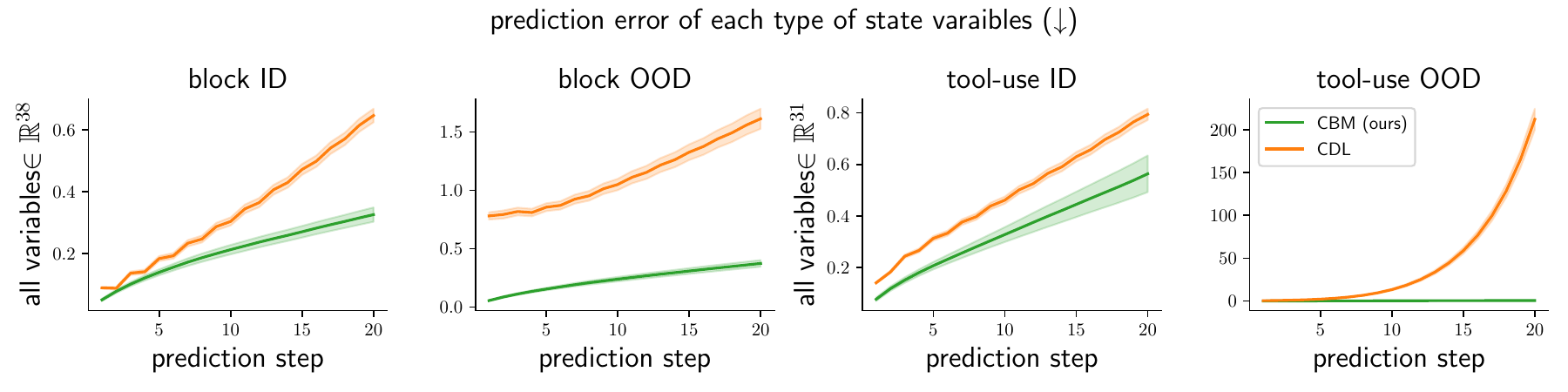}
\caption{Dynamics prediction error in the block and tool-use environments.}
\label{fig:dyn_pred_overall}
\end{figure}

We also evaluate CBM using implicit dynamics and CDL in terms of prediction accuracy. Specifically, given $s_t$ and $a_{t:t+19}$, we use them to generate 20-step predictions, i.e., $s_{t+1:t+20}$, on both in-distribution (ID) and out-of-distribution (OOD) $s_t$ for the block and tool-use environments. For OOD states, distractor values are replaced with random values sampled from $\mathcal{N}(0, 100)$. The results are again measured on 10K transitions for each method. 

As shown in Fig. \ref{fig:dyn_pred_overall}, when measuring the prediction error of all state variables, CBM has lower prediction error than CDL on both ID and OOD states. Especially on OOD states, as CBM learns fewer spurious correlations than CDL, it keeps similar performance while CDL's errors increase significantly compared to ID states.

\subsection{Dynamics Computation Cost}
When predicting each single state variable, though our method draw 8192 samples and compute their energies, we want to point out that the computation cost is still comparable to explicit dynamics models and does not prevent the implicit dynamics from scaling to environments with large state spaces.

Requiring a large number of samples is an inherent weakness of implicit models. Nevertheless, this weakness has not prevented applying a similar model to robotic tasks (Florence et al., 2022) and to higher-dimensional states such as images (Chen et al., 2020b).

Meanwhile, the computational complexity is $O(d_S^2)$ for explicit dynamics and $O(d_S \times (d_S + N))$ for implicit dynamics (where $d_S$ is the number of state variables and $N$ is the number of samples). So for higher $d_S$, the complexity ratio of implicit over explicit dynamics ($=1+\frac{N}{d_S}$) is actually lower.

Moreover, for small $d_S$ where the complexity ratio is high, one can trade off between computation and prediction accuracy. We show the mean and standard deviation of prediction performance (measured as one-step prediction error on all state variables) and wall time (in seconds) when choosing different numbers of samples and compare them with explicit models in the block environment ($d_S=47$) on 25K transitions.

\begin{table}
\centering
\caption{Prediction error and wall time for explicit dynamics and implicit dynamics with different number of samples.}
\begin{small}
\begin{tabular}{cccccccc}
\toprule
& \multirow{2}{*}{explicit}             & \multicolumn{6}{c}{implcit (ours)} \\
&                                       & 256   & 512  & 10  & 2048    & 4096    & 8192 \\
\midrule
prediction error    & \thead{0.09 $\pm$ 0.05}  & \thead{0.17 $\pm$ 0.01}  & \thead{0.10 $\pm$ 0.01}   & \thead{0.07 $\pm$ 0.00}    & \thead{0.06 $\pm$ 0.00} & \thead{\textbf{0.05} $\pm$ 0.00}  & \thead{\textbf{0.05} $\pm$ 0.00} \\
wall time (s)       & \thead{\textbf{5.79} $\pm$ 0.21}  & \thead{11.33 $\pm$ 0.06}  & \thead{16.47 $\pm$ 0.40}  & \thead{25.69 $\pm$ 0.37}  & \thead{47.86 $\pm$ 0.19}  & \thead{83.47 $\pm$ 2.44}  & \thead{163.35 $\pm$ 5.35}\\
\bottomrule
\end{tabular}
\end{small}
\vspace{-0pt}
\label{tab:dyn_pred_cost}
\end{table}

As shown in Table \ref{tab:dyn_pred_cost}, one can use fewer samples (e.g. 1024 or 2048) to achieve a prediction error similar to using 8192 samples. Notice that implicit dynamics with those smaller sample sizes take only about 5x or 8x of the computation time for explicit dynamics, much lower than the theoretical ratios (i.e., $1+\frac{N}{d_S}$ which are around 23x or 45x respectively), as we use a smaller network to extract feature from samples $s^{i, n}_{t+1}$ than from the current state variable $s^j_t$.

\section{Task Learning Implementation and Further Results}
\label{app:task_results}
In this section, we give more details on the RoboSuite environments (block and tool-use) used in the main paper, and methods for the sample efficiency experiments. 

\subsection{Task Rewards}

For DMC tasks, we use the reward function in the official implementation code. 
For Robosuite tasks, let $eef_t \in \mathbb{R}^3$ be the current end-effector position and $g \in \mathbb{R}^3$ is the target position in this episode. 
The reward functions for the tasks used in the experiments of Sec. \ref{exp:task}, are defined as follows:

\textbf{Pick} (B): raise the block $mov$ to the target position $g$,
\begin{align*}
r_t = 0.2  &(1 - \tanh{(2.0 || eef_t - mov_t ||_2})) \\
+& \mathds{1}\left[\text{$mov$ is grasped}\right] (0.4 + 0.5 (1  - \tanh{(5.0 || mov_t - g||_2)}) \\
+& \mathds{1}\left[||mov_t - g||_2 < 0.05 \right].
\end{align*}
\textbf{Stack} (B): stack the movable object $mov$ on the top of the unmovable object $unm$.
\begin{align*}
r_t = 0.2 \cdot &(1 - \tanh{(2.0\lVert eef_t - mov_t\rVert_2))} \\ 
+& 0.4 \cdot \mathds{1}\left[\text{$mov$ is grasped}\right] \\
+&  0.5 (1 - \tanh{(5 ||mov_{x,y} - unm_{x, y}||_1)})\cdot \mathds{1}\left[ mov0 \text{ is lifted}\right] \\
+& 2.0 \cdot \mathds{1}\left[success\right], 
\end{align*}
where the notation $mov_{x, y}$ refers to the $x$ and $y$ coordinates of $mov$, and similarly for $unm$.

\textbf{Series} (T):  use the L-shaped tool to move the faraway block closer to the robot, then pick up the block and place it in the pot.
\begin{align*}
r_t = 0.2 &\cdot (1 - \tanh{(2.0\lVert eef_t - tool_t\rVert_2))} \\ 
+& 0.2 \cdot \mathds{1}\left[\text{$tool$ is grasped}\right] \\
+& 0.2 \cdot  (1 - \tanh{(5 \lVert tool_t - block_t\rVert_2)})\cdot \mathds{1}\left[ tool \text{ is grasped}\right] \\
+& 0.2 \cdot  (1 - \tanh{(5 \max(-block_x, 0)})\cdot \mathds{1}\left[ tool \text{ is grasped}\right] \\
+& 0.4 \cdot \mathds{1}\left[\text{$tool$ is not grasped}\right] \cdot \mathds{1}\left[ block_x < 0 \right] \\
+& 0.4 \cdot  (1 - \tanh{(5 \lVert eef_t - tool_t\rVert_2})\cdot \mathds{1}\left[ block_x < 0 \right] \\
+& 0.6 \cdot \mathds{1}\left[\text{$block$ is grasped}\right] \\
+& 0.5 \cdot  (1 - \tanh{(5 \lVert block_{x, y} - pot_{x, y}\rVert_2)})\cdot \mathds{1}\left[ block \text{ is grasped}\right] \\
+& 2.0 \cdot \mathds{1}\left[success\right]. 
\end{align*}

\subsection{State Abstraction Learning Methods}

\begin{table}
\centering
\caption{Parameters of the reward predictor and SAC. Parameters shared if not specified.}
\begin{small}
\begin{tabular}{ccccccc}
\toprule
\textbf{Method} & \textbf{Name}                 & \multicolumn{5}{c}{\textbf{Tasks}} \\
                &                               & \thead{Pick (B)}  & Stack (B) & Series (T) & Cheetah (T)  & Walker (DMC)   \\
\midrule
\multirow{6}{*}{\thead{Reward\\Predictor}}
& feature architecture              & \multicolumn{5}{c}{[128, 128]} \\
& predictor architecture            & \multicolumn{5}{c}{[128, 128]} \\
& activation functions              & \multicolumn{5}{c}{ReLU}     \\
& training step                     & \multicolumn{5}{c}{50K}    \\
& learning rate                     & \multicolumn{5}{c}{3e-4}                              \\
& batch size                        & \multicolumn{5}{c}{64}                            \\
\midrule
\multirow{16}{*}{SAC}
& horizon                           & \multicolumn{2}{c}{250}     &\multicolumn{1}{c}{400}     &\multicolumn{2}{c}{1000}        \\
& actor architecture                & \multicolumn{5}{c}{[256, 256]}                             \\
& critic architecture               & \multicolumn{5}{c}{[256, 256]}                             \\
& actor activation functions        & \multicolumn{5}{c}{[Relu, Relu]}                           \\
& critic activation functions       & \multicolumn{5}{c}{[Relu, Relu]}                           \\
& TD steps                          & \multicolumn{5}{c}{1}                                      \\
& batch size                        & \multicolumn{5}{c}{256}                                  \\
& grad clip norm                    & \multicolumn{5}{c}{10}                                    \\
& actor/critic learning rate        & \multicolumn{5}{c}{1e-4}                                  \\
& tau                               & \multicolumn{5}{c}{5e-3}                                  \\
& gamma                             & \multicolumn{5}{c}{0.99}                                  \\
& buffer size                       & \multicolumn{5}{c}{5e6}                                  \\
& alpha start                       & 0.9     & 0.9    & 0.9  & \multicolumn{2}{c}{0.5}                      \\
& alpha finish                      & 0.1     & 0.05   & 0.1  & \multicolumn{2}{c}{0.1}                        \\
& alpha decay                       & 0.666   & 3.333  & 5    & \multicolumn{2}{c}{1}                          \\
\bottomrule
\end{tabular}
\end{small}
\vspace{-0pt}
\label{tab:task_policy_parameters}
\end{table}
For Pick and Stack, the dynamics model is pretrained for all methods and frozen during task learning, and only the reward predictor is learned. For Series and DMC tasks, the dynamics model is learned from scratch during task learning, jointly with the policy and the reward predictor.

\paragraph{CDL}
For CDL's task-independent abstraction,  the dynamics model is trained offline on 2M pre-collected transitions using scripted policies and then used to derive the abstraction, which remains fixed during task learning. 

\paragraph{TIA and Denoised MDP}
They are originally implemented for representation learning of image state spaces. To adapt to factored state spaces, we replace their encoders and decoders with a binary mask that partitions state variables into task-relevant or irrelevant parts. The mask is jointly optimized with the dynamics and reward models using the Gumbel reparameterization trick \citep{jang2016categorical}. 
While the original implementations of TIA and Denoised MDP both learn dynamics from scratch during task learning, in Pick and Stack tasks, CDL and CBM use a task-independent dynamics model that is already trained by the task-learning phase. Thus, for a fair comparison, TIA and Denoised MDP are also initialized with pretrained dynamics models, where only the final layer of the model is allowed to vary during task learning. We also tried training the dynamics model from scratch, but this led to worse performance.
For TIA and Denoised MDP, we conducted a search on the following hyperparameters:
\begin{itemize}
    \item Gumbel temperature scheduling, among the final temperature of $\{0.05, 0.1, 0.2, 0.5, 1.0, 2.0, 5.0, 10.0\}$.
    \item Regularization coefficient on size of the abstraction, i.e., the number of the reward-relevant state variables for TIA, and the number of controllable and reward-relevant variables for Denoised MDP, among $\{10^{-3}, 10^{-4}, \cdots, 10^{-7}\}$.
\end{itemize}

We hypothesize that the reason why TIA and Denoised MDP perform worse compared to CBM (Sec \ref{exp:task}) is that both methods rely on assumptions that do not hold in general MDPs, leading to inaccurately learned abstractions and inefficient task learning. Specifically, TIA assumes there is no dynamical dependency between two segments in the representation space (i.e., state variable space in our setting) — reward-relevant one and reward-irrelevant one. Meanwhile, Denoised MDP makes a similar assumption about dynamical independence between different segments. However, in general MDPs, reward-irrelevant state variables could be dynamically dependent on reward-relevant variables. For example, when the task is to pick block A with the robot arm, though block B is reward-irrelevant, block B still can be manipulated by the arm. As a result, block B will also be included in the same segment to minimize its prediction error. Hence, TIA and Denoised MDP often include all controllable variables in their abstractions. In contrast, CBM doesn’t have this “dynamic independence” assumption and identifies dynamical causal relationships between state variables, thus it can remove controllable variables that are task-irrelevant from its abstraction.

\paragraph{SAC Policy}
We adopt the implementation of SAC by Tianshou \citep{tianshou}. For most SAC parameters, tuned  hyperparameter values published by prior literature worked well across all tasks. The exception is $\alpha$, the entropy  regularization coefficient, which controls the relative weights of the return and entropy terms in SAC. We found that automatic entropy scheduling often led to early convergence and task failure, so for each task, we defined a manual entropy decay schedule, which we tuned separately for each task. The entropy schedule is a function of both the current and total training steps (a constant). It is defined below and has the following parameters: $\alpha_{start}, \alpha_{finish}, \alpha_{decay}$.

\begin{equation*}
\alpha(t)_{t_{total}} = (\alpha_{start} - \alpha_{finish}) \exp{\left( \frac{-\alpha_{decay} \cdot t}{t_{total}}\right)} + \alpha_{finish}.
\end{equation*}

The architecture and hyperparameters of SAC are listed in Table \ref{tab:task_policy_parameters}. SAC hyperparameters are shared across all methods for each task. We select between [0.5, 1.0] for $\alpha_{start}$, [0.0, 0.2] for $\alpha_{finish}$, and [0.5, 5] for $\alpha_{decay}$ for best SAC performance.

\paragraph{CBM}
For CBM, the reward predictor is jointly trained with the policy, the learned reward causal graph may change during training and thus change the derived bisimulation abstraction. When the abstraction changes, CBM resets $\pi$ and relearns the policy from existing data in the replay buffer. Policy resetting is a technique proposed by \citet{pmlr-v162-nikishin22a}, who showed that periodically resetting the policy and retraining from the replay buffer may improve both sample efficiency and asymptotic performance for deep RL agents. For a fair comparison, policy resetting is applied to TIA and Denoised MDP as well. The architecture and hyperparameters of the reward predictor are listed in Table \ref{tab:task_policy_parameters}.

\subsection{Learned Task-Specific Abstraction}
\label{app:abstraction}

\begin{figure}[hbtp]
  \centering
  \includegraphics[width=1.0\linewidth]{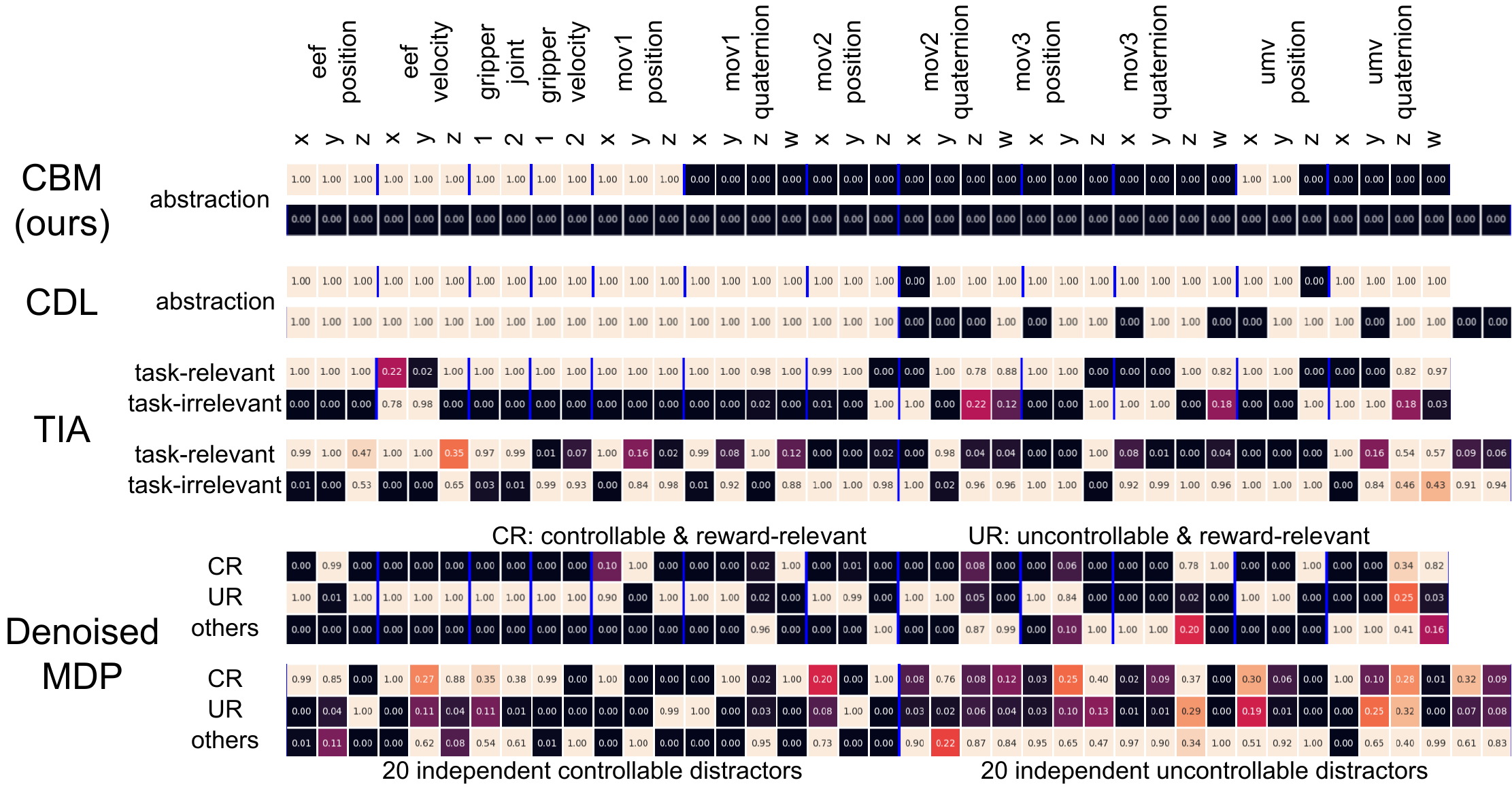}
  \caption{\small State abstractions learned by CBM, TIA, and Denoised MDP for the Stack task.}
  \label{fig:variable_level_abstraction}
  \vspace{-0pt}
\end{figure}

For the Stack task, the learned state abstraction by each method is shown in Fig. \ref{fig:variable_level_abstraction}. Again, CBM keeps all reward-influencing variables and their causal ancestors. Despite our efforts to hyperparameter tune, TIA and Denoised fail to learn meaningful abstractions. We found that both methods were highly sensitive to their regularization coefficients. Note that the Stack task also violates their assumptions of each component having independent dynamics, which may explain their failure to learn good abstractions. 

\subsection{Task Learning Ablation}
\begin{figure}[ht]
  \centering
  \includegraphics[width=0.5\columnwidth]{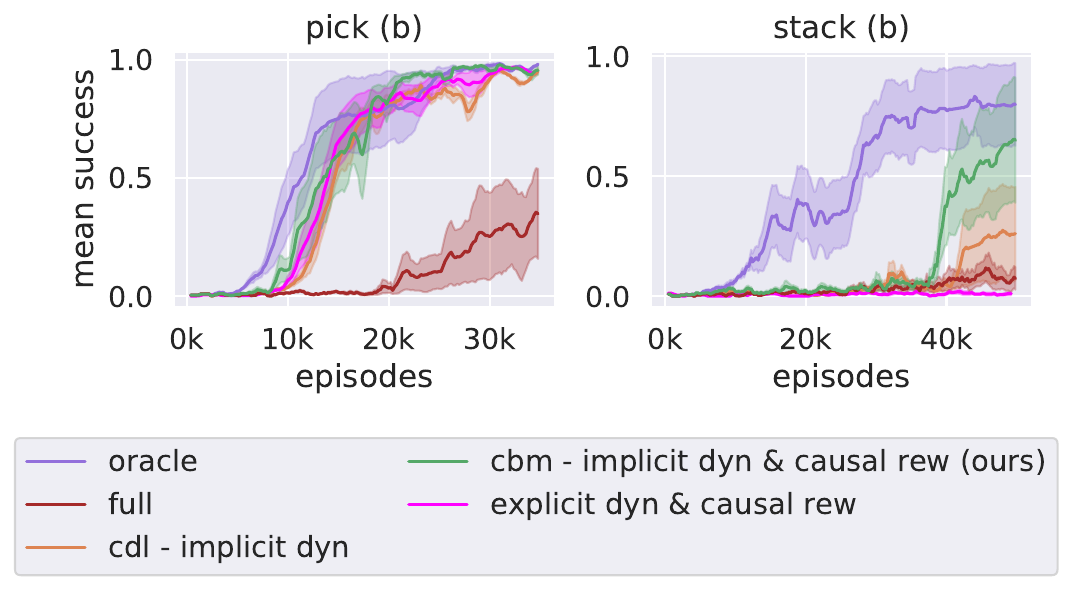}
  \caption{\small Learning curves of CBM and CDL (which both use implicit dynamics in the main paper), and an ablation of CBM that uses explicit dynamics on Pick and Stack tasks. We observe that the ablation has much worse sample efficiency on Stack.}
  \label{fig:cdl_explicit_sample_eff}
\end{figure}

CBM learns implicit dynamics and a causal reward function. In Sec. \ref{exp:dynamics}, we show that implicit dynamics models surpass explicit models in terms of causal graph accuracy and state abstraction accuracy. Fig. \ref{fig:cdl_explicit_sample_eff} shows an ablation of CBM which instead uses explicit dynamics. We observe that on the Pick task, the difference from CBM is not significant as Pick is relatively easy.
On the more challenging Stack task where accurate abstraction plays a more important role, the performance of explicit dynamics is much worse than CBM which uses implicit dynamics.

\section{Compute Architecture}
\label{app:compute_architecture}

The code is implemented with pyTorch. The 5 seeds selected are 0 - 4, and the seed can be specified in the configuration file.
The experiments were conducted on machines of the following configurations: 
\begin{itemize}
    \item Nvidia Titan V GPU; Intel(R) Xeon(R) CPU E5-2630 v4 @2.20GHz
    \item Nvidia V100-SXM2 GPU; Intel(R) Xeon(R) CPU E5-2698 v4 @2.20GHz
    \item Nvidia A40 GPU; Intel(R) Xeon(R) Gold 6342 CPU @2.80GHz
    \item Nvidia A100 GPU; Intel(R) Xeon(R) Gold 6342 CPU @2.80GHz
\end{itemize}